%% file: main.tex
\documentclass[10pt,twocolumn,letterpaper]{article}

\usepackage[pagenumbers]{cvpr} 
\usepackage{multirow}
\input{preamble}

\definecolor{cvprblue}{rgb}{0.21,0.49,0.74}
\usepackage[pagebackref,breaklinks,colorlinks,allcolors=cvprblue]{hyperref}

\def\confName{CVPR}
\def\confYear{2026}

\title{VG-Refiner: Towards Tool-Refined Referring Grounded  Reasoning \\ via Agentic Reinforcement Learning}

\author{Yuji Wang$^{1, 2}$, Wenlong Liu$^2$, Jingxuan Niu$^1$, Haoji Zhang$^1$, Yansong Tang$^1$\footnotemark[2]~\\
$^1$Tsinghua Shenzhen International Graduate School, Tsinghua University \\$^2$International Digital Economy Academy (IDEA)\\
}

\begin{document}
\twocolumn[{%
    \renewcommand\twocolumn[1][]{#1}%
    \maketitle
   
    \vspace{-18pt}
    \begin{center}
        \captionsetup{type=figure}
        \centering
        \includegraphics[width=1.0\linewidth]{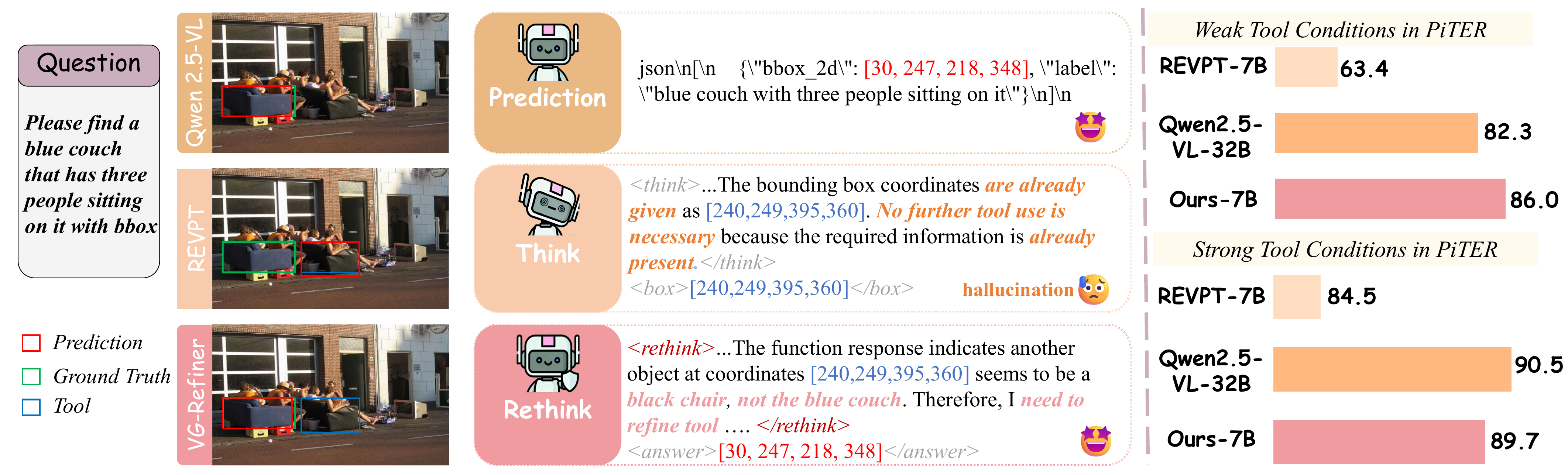}
        \captionof{figure}{In the left case, \textbf{VG-Refiner} performs explicit reasoning over the tool outputs via the CoT process, whereas \textbf{REVPT} merely confirms the tool feedback without any analytical examination, leading to its inability to detect tool-induced errors. The baseline model Qwen2.5-VL-7B of REVPT and VG-Refiner can locate the true object by its own capabilities without CoT. The right part shows that our \textbf{VG-Refiner} achieves grounding accuracy comparable to the 32B model across the average of five test splits on the RefCOCO series, under different tool conditions in the PiTER evaluation protocol.}
        \label{teaser}
    \end{center}%
    
}]

\setcounter{footnote}{0}
\renewcommand{\thefootnote}{}
\setcounter{footnote}{1}
\renewcommand{\thefootnote}{\fnsymbol{footnote}}
\footnotetext[2]{Corresponding author}
\setcounter{footnote}{0}
\renewcommand{\thefootnote}{\arabic{footnote}}

\input{sec/0_abstract}    
\input{sec/1_intro}
\input{sec/2_relatedwork}
\input{sec/3_method}

\input{sec/4_experiment}

\input{sec/5_conclusion}

{
    \small
    \bibliographystyle{ieeenat_fullname}
    \bibliography{main}
}

\input{sec/X_suppl}

\end{document}

%% file: preamble.tex
\usepackage{multirow}
\usepackage{colortbl}
\usepackage{amsmath}

%% file: sec/0_abstract.tex
\begin{abstract}
Tool-integrated visual reasoning (TiVR) has demonstrated great potential in enhancing multimodal problem-solving. However, existing TiVR paradigms mainly focus on integrating various visual tools through reinforcement learning, while neglecting to design effective response mechanisms for handling unreliable or erroneous tool outputs. This limitation is particularly pronounced in referring and grounding tasks, where inaccurate detection tool predictions often mislead TiVR models into generating hallucinated reasoning. To address this issue, we propose the \textbf{VG-Refiner}, the first framework aiming at the tool-refined referring grounded reasoning. Technically, we introduce a two-stage think–rethink mechanism that enables the model to explicitly analyze and respond to tool feedback, along with a refinement reward that encourages effective correction in response to poor tool results. In addition, we propose two new metrics and establish fair evaluation protocols to systematically measure the refinement ability of current models. We adopt a small amount of task-specific data to enhance the refinement capability of VG-Refiner, achieving a significant improvement in accuracy and correction ability on referring and reasoning grounding benchmarks while preserving the general capabilities of the pretrained model. Our code is available at https://github.com/VoyageWang/VG-Refiner.

\end{abstract}

%% file: sec/1_intro.tex
\section{Introduction}
Tool-integrated visual reasoning (TiVR) can significantly improve the performance of large vision-language models (LVLMs) across diverse specialized domains, such as table or document question-answering (QA) \cite{dianjinocr, openthinkimg, vtool-r1}, visual search and reasoning \cite{deepeyes,mini_o3}, and general visual QA \cite{search-r1,thinking-with-videos,ponder}. Building upon recent advances that incorporate verbal chain-of-thought (CoT) reasoning to strengthen LVLMs \cite{deepseek-r1,visual-rft,univg,seg-zero}, existing approaches often employ reinforcement learning (RL) to train models for tool-based reasoning tasks \cite{visual-arft,VILASR,thyme,revpt}. This training paradigm not only reduces the need for extensive manual annotations but also enables the models to achieve stronger generalization capabilities across diverse tasks \cite{survey-agent}.

However, this agentic RL paradigm focuses on teaching the model how to select suitable tools from a fixed toolkit, such as detection \cite{grounding-dino,grounding-dino1.5}, depth \cite{depth-anything,depth-anythingv2}, and edge tools, which overlooks situations where the selected tool's outputs are unreliable or incorrect. The existing TiVR model can be easily disturbed by this poor feedback and then generate a hallucinatory response to explain and reasoning based on it. This issue is particularly prominent in referring expression comprehension (REC) tasks \cite{rec-survey}, where detection tools frequently produce inaccurate or misleading results. Thus, it further limits the applications in downstream tasks, such as spatial grounding ~ \cite{lavt,iterprime,lisa,llava-grounding} or counting \cite{referring-counting}.

To be more specific, as shown in Figure \ref{teaser}, we require the TiVR model REVPT \cite{revpt}, to locate the referred object and prompt it with the incorrect tool results. Despite the explicit verbal CoT process, the model fails to discover the irrationality of the prompted tool result, even though the original Qwen2.5-VL-7B model \cite{qwen2.5VL} can locate the true object. We attribute this scenario to the notable deficiency in their ability to discover and correct tool-induced errors.

To overcome this limitation, we propose a tool-refined referring grounded reasoning (TrRGR) framework, \textbf{VG-Refiner}, that can rethink the feedback from the expert REC tools and decide whether to accept or refine the results through explicit analysis trained based on agentic RL. Rather than relying on tools as infallible sources of information, VG-Refiner performs a secondary reasoning stage to analyze the contextual consistency between the visual evidence, the textual query, and the tool-generated predictions. Although VG-Refiner is primarily designed to address unreliable or erroneous tool outputs, it also demonstrates a strong follow-correct rate when tools are reliable.

Technically, we propose a novel think–rethink two-stage refinement framework, where the model first performs independent reasoning and then re-evaluates its conclusions by integrating feedback from external tools to produce more reliable and accurate results. Secondly, we design a novel refinement reward to encourage the model to recognize and accept reliable feedback from external tools while reinforcing its ability to detect and refine erroneous predictions. In addition, to comprehensively evaluate the refinement ability of different models, we establish a prompt-integrated tool enhancement and refinement (PiTER) evaluation protocol and create two tool-refined metrics that assess model performance from two key dimensions: refinement capability and refinement quality. The PiTER protocol injects the tool results into the system prompt and outputs with JSON grounding results fairly. Overall, our contributions are summarized as follows:
\begin{itemize}
\item We identify a key limitation of existing TiVR methods in REC tasks and propose \textbf{VG-Refiner}, a two-stage think–rethink framework equipped with a refinement reward that enhances both reasoning and correction while preserving trust in reliable tool feedback.

\item We define the TrRGR reasoning paradigm and propose two novel metrics for assessing the refinement capability of all LVLMs, and establish a unified PiTER protocol to ensure fair comparison.

\item Preserving the general capabilities of the pre-trained model, our model outperforms SOTA methods on RefCOCO/+/g in accuracy and surpasses the Qwen2.5-VL-32B model in refinement ability, shown in Figure \ref{teaser}, while trained on a small amount of task-specific data.
\end{itemize}


%% file: sec/2_relatedwork.tex
\section{Related Work}

\begin{figure*}[ht]
	\centering	\includegraphics[width=\textwidth]{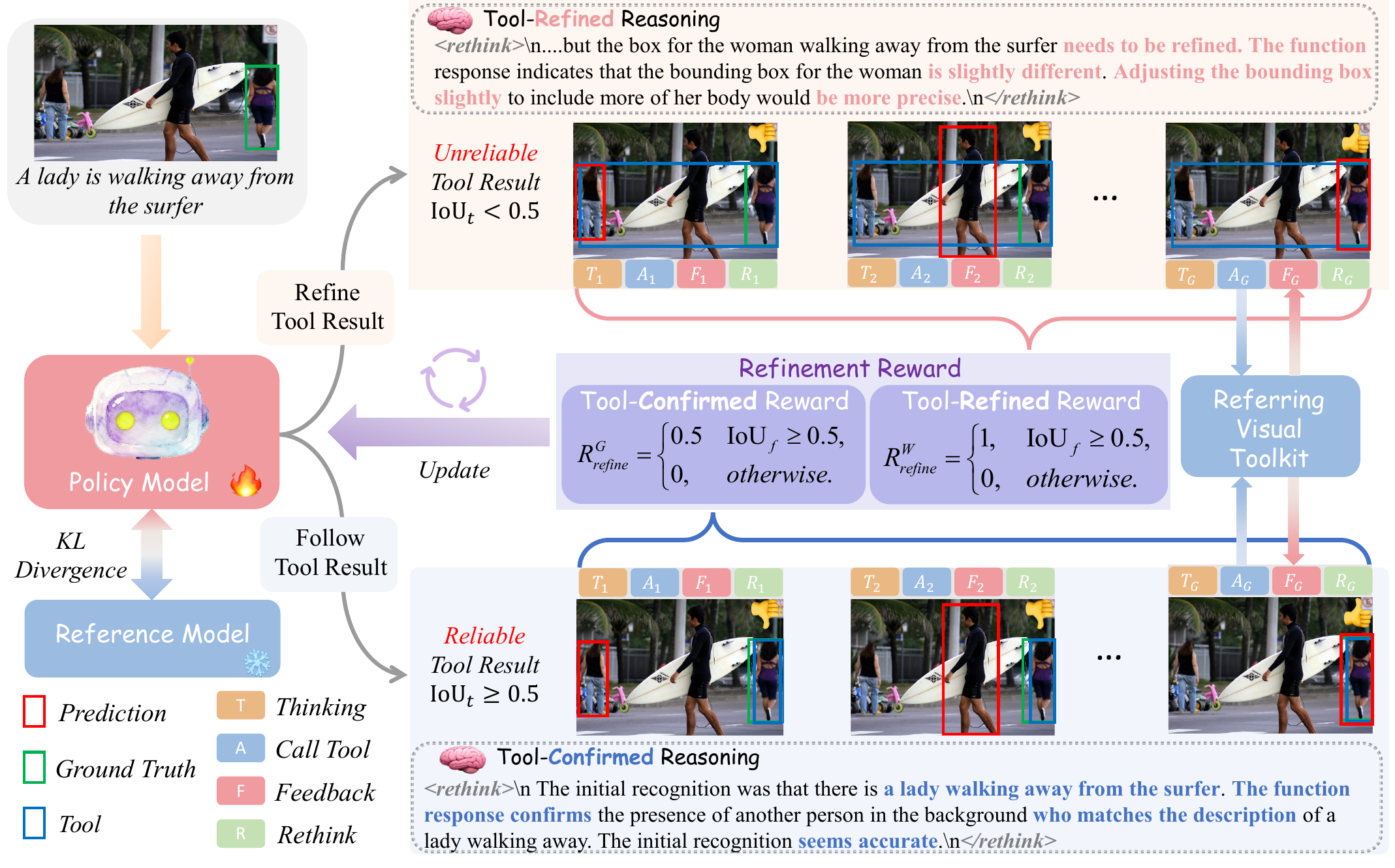}
	\caption{The overall framework of \textbf{VG-Refiner}. In our reward design, we consider the quality of tool feedback $\text{IoU}_t$. For different circumstances, we adopt different levels of reward to encourage the model to refine the tool's incorrect results or accept the reliable results. We use GRPO to optimize the policy model, which produces various $G$ rollouts during training. After the {think} process, the model queries a referring visual toolkit for additional reference outputs. In GRPO, KL divergence constrains strategy deviation from the frozen reference model to ensure stable optimization.}
	\label{framework}
\end{figure*} 


\textbf{Tool-integrated Visual Reasoning.} Tool-integrated visual reasoning (TiVR) significantly improves the reasoning abilities of LVLMs. The effectiveness of TiVR hinges on well-curated and diverse tools, which can be categorized into three groups \cite{thinking_withimage_survey}. Firstly, to mimic human visual attention, current works, including DeepEyes \cite{deepeyes}, Mini-O3 \cite{mini_o3}, Simple-o3 \cite{simple_o3}, and Pixel Reasoner \cite{pixel_reasoner} focus on iterative zoom-in and region-of-interest (RoI) selection, equipping LVLMs with active perception capabilities. The second group includes OCR \cite{dianjin-ocr} and code interpretation tools \cite{visual-arft, vtool-r1}, which empower LVLMs to extract textual information from images and perform numerical or logical reasoning through executable Python code. Finally, REVPT \cite{revpt}, OpenThinkIMG \cite{openthinkimg}, and Active-O3 \cite{active_o3} employ visual perception tools, such as depth estimation and detection modules, to enhance perception capabilities, e.g., counting. However, without carefully designed refinement mechanisms, these models are easily disturbed by the imperfect or noisy outputs of detection tools in REC.

\textbf{LVLMs for REC Tasks.} The referring expression comprehension aims to locate the visual element aligned with the linguistic input phrase, which has many applications across diverse tasks \cite{sam2-love,flash-vstream,rex_omni,iterprime,unilseg,soc,lavt}. With the rapid development of LVLMs \cite{llava,qwen2,qwen3_omni,lisa}, many series of works have used the LVLMs for REC tasks. LLAVA-grounding \cite{llava-grounding} and Grounding-GPT \cite{groundinggpt} construct large-scale datasets using the supervised fine-tuning (SFT) method to enhance grounding capabilities. Recently, studies like VLM-R1 \cite{vlm-r1}, Visual-RFT \cite{visual-rft}, Visionreasoner \cite{visionreasoner}, and Ground-R1 \cite{ground-r1} have demonstrated that the RL can make stronger grounding reasoning abilities than SFT with verbal CoT process. Rex-Thinker \cite{rex_thinker} and Rex-Seek \cite{rex_seek} are also trained with RL and employ a proposal detection tool, upon which the LVLM performs reasoning over the proposed bounding boxes to retrieve the true target object. Therefore, no existing methods employ agentic RL for the REC with respect to external references provided by tools.

\textbf{Expert REC Models.} We define a good expert REC model as one that is fine-tuned on referring datasets such as RefCOCO, RefCOCO+, and RefCOCOg \cite{refcoco}, while a weak model refers to one that possesses detection capabilities but has not been fine-tuned on any referring dataset. Current state-of-the-art expert models, such as EVF-SAM \cite{evf-sam} and InstanceVG \cite{InstanceVG}, employ bi-directional encoders BEIT-3 \cite{beit_3} to achieve effective language–vision fusion.

In summary, existing methods overlook the issue of erroneous tool outputs in REC scenarios. Therefore, we propose the tool-refined referring grounded reasoning paradigm.

%% file: sec/3_method.tex
\section{Method}

In this section, we present a comprehensive description of our proposed \textbf{VG-Refiner}. We begin by introducing the special features of TrRGR and the reasoning paradigm, followed by a detailed explanation of its training strategy. Finally, we describe the evaluation protocols for assessing refinement capabilities and the corresponding metrics.

\subsection{Tool-refined Referring Grounded Reasoning}
\textbf{Concept Constructions.} We propose tool-refined referring grounded reasoning (TrRGR), a new paradigm for REC tasks. We differentiate our approach from two prevailing paradigms: (1) conventional TiVR methods \cite{revpt,openthinkimg} that merely integrate detection tool invocation and feedback without deeper reasoning against unreliable tool results, and (2) verbal CoT-based models \cite{seg-zero,ground-r1,univg} that rely solely on verbal CoT reasoning without external tool collaboration. In contrast, built upon internal reasoning, TrRGR introduces an additional explicit rechecking and refinement mechanism. Through a structured CoT process, the model analyzes and verifies the outputs of external expert tools, enabling it to reason about and correct unreliable results. Moreover, it optimizes reliable tool outputs to achieve higher Intersection-over-Union (IoU) scores. 



\begin{figure*}[ht]
	\centering	\includegraphics[width=0.9\textwidth]{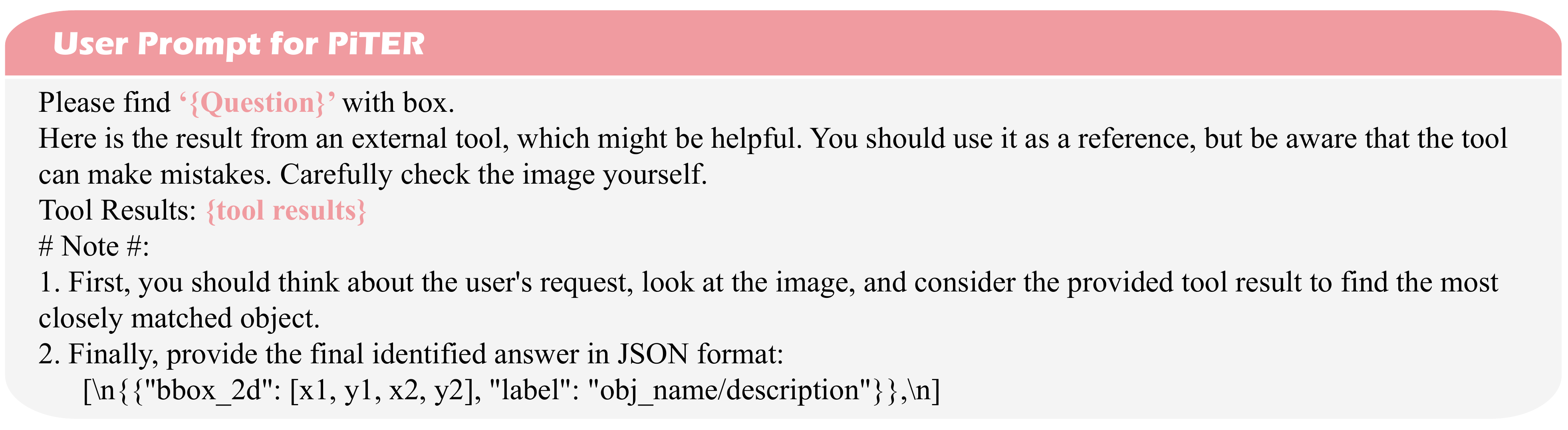}
	\caption{User prompt for the PiTER evaluation process. This prompt is shared across all model types, requiring the model to produce grounding results in a JSON format through a single-stage conversation, without any CoT reasoning or tool interaction. The placeholder $\{\text{Question}\}$ is replaced with the referring expression, while $\{\text{tool results}\}$ is substituted with the feedback from either a strong or weak tool corresponding to the given question.}
	\label{user_prompt}
\end{figure*} 

\textbf{Reasoning Paradigm.} The overall architecture of our VG-Refiner is shown in Figure \ref{framework}. We employ Qwen2.5-VL-7B as the refiner policy model due to its strong general reasoning capabilities demonstrated by \cite{revpt,seg-zero,visionreasoner}. To enhance reasoning reliability under uncertain tool predictions, we design a two-stage think–rethink framework that mimics the human cognitive process from recognizing, verifying the tool feedback, to correcting. In the first-round {think} stage, the policy model $\pi_m$ first take the input image $I$ and referring phrase $P_q$ with the system prompt $P_\text{sys}$ and output the thinking CoT reasoning $P_\text{CoT}$, which is included between the \textless \textit{think}\textgreater ~and \textless \textit{/think}\textgreater ~and the tool calling action $P_A$, shown below:
\begin{equation}
    P^1_\text{CoT}, P_A = \pi^1_{m}(I, P_{q}, P_{\text{sys}}),
\end{equation}
where $\pi^1_{m}$ refers to the first conversation. This stage aims for broad coverage rather than precision, providing an initial hypothesis of the referred object. The action phrase will be parsed and used for querying external tool results. Next, the expert model produces its detection result $B_{{t}}$ based on $I$ and $T$, which serves as external feedback for the refinement process. In the second-round {rethink} stage, the model re-evaluates both initial predictions and performs targeted correction analysis included between \textless \textit{rethink}\textgreater~and \textless \textit{/rethink}\textgreater, shown below:
\begin{equation}
    B_f, P^2_\text{CoT} = \pi^2_{m}(I, P^1_\text{CoT}, B_{t}),
\end{equation}
where $B_f$ is the final refined result produced by the policy model. Through this reflective process, the refiner learns when to trust or override the expert model. This iterative reasoning effectively combines tool-assisted perception with self-corrective reflection, leading to more accurate and interpretable REC outcomes.

\textbf{Tool Model.} In the REC task, we define a \textit{strong tool} as one that has already achieved state-of-the-art (SOTA) performance on mainstream REC benchmarks. In contrast, a \textit{weak tool} refers to a model that has not been fine-tuned on any referring expression dataset. During inference, we employ the strong tool to achieve superior performance, surpassing both the tool itself and the base MLLM. However, when benchmarking the model’s refinement capability, we adopt the weak tool to better assess how effectively the model can correct suboptimal tool predictions. Each referring visual tool takes the same inputs $(I, P_q)$ and outputs its predicted bounding boxes $B_t$.


\begin{table*}[t]
    \centering
    \caption{Evaluation results of visual grounding in $Acc@0.5$ on RefCOCO, RefCOCO+, and RefCOCOg datasets. }
    \label{tabt1}
    \begin{tabular*}{\textwidth}{@{\extracolsep{\fill}}lcccccccccc@{}} 
        \toprule
        \multirow{2}{*}{Method} 
        & \multicolumn{3}{c}{RefCOCO} 
        & \multicolumn{3}{c}{RefCOCO+} 
        & \multicolumn{2}{c}{RefCOCOg} 
        & \multirow{2}{*}{Avg.} \\ 
        \cmidrule(lr){2-4} \cmidrule(lr){5-7} \cmidrule(lr){8-9}
        & val & testA & testB & val & testA & testB & val & test &  \\ 
        \midrule 
        Qwen2.5-VL-7B \cite{qwen2.5VL} & 90.0 & 92.5 & 85.4 & 84.2 & 89.1 & 76.9 & 87.2 & 87.2 & 86.6 \\
        Qwen2.5-VL-72B \cite{qwen2.5VL} & 92.7 & \underline{94.6} & 89.7 & \textbf{88.9} & 92.2 & \textbf{83.7} & \underline{89.9} & {90.3} & \underline{90.3} \\
        Ground-R1 \cite{ground-r1} & \underline{92.9} & 93.9 & 88.0 & 86.5 & 90.8 & 78.8 & \textbf{90.1} & 90.2 & 88.9 \\
        CogVLM \cite{cogvlm} & 92.5 & 94.0 & 88.7 & 87.5 & 91.8 & 81.4 & 89.5 & 90.1 & 89.4 \\
        CogCoM \cite{cogcom} & 92.3 & \underline{94.6} & 89.2 & 88.2 & \textbf{92.8} & 82.1 & 89.3 & \underline{90.5} & 89.9 \\
        UniVG-R1 \cite{univg} & 91.6 & 93.1 & 87.2 & 85.9 & 90.5 & 80.0 & 88.7 & 88.6 & 88.2 \\
        Vitron \cite{vitron} & 90.9 & 93.2 & 89.3 & 83.7 & 89.1 & 76.9 & 86.4 & 87.0 & 87.1 \\
        UniPixel \cite{unipixel} & 92.0 & 92.5 & 88.1 & 87.2 & 91.9 & 82.1 & 88.6 & 88.7 & 88.9 \\
        \hline
        EVF-SAM (tool) \cite{evf-sam} & 92.5 & 94.2 & \underline{90.3} & 86.4 & 90.2 & 81.7 & 87.7 & 88.9 & 88.9 \\
        {VG-Refiner (Ours)} & \textbf{93.2} & \textbf{95.0} & \textbf{90.7} & \underline{88.5} & \underline{92.7} & \underline{83.0} & {89.8} & \textbf{90.6} & \textbf{90.5} \\
        \rowcolor{gray!10}
        $\Delta$ vs. EVF-SAM (tool) 
        & +0.7 & +0.8 & +0.4 
        & +2.1 & +2.5 & +1.3 
        & +2.1 & +1.7 & +1.6 \\
        \rowcolor{gray!10}
        $\Delta$ vs. Qwen2.5-VL-7B 
        & +3.2 & +2.5 & +5.3 
        & +4.3 & +3.6 & +6.1 
        & +2.6 & +3.4 & +3.9 \\
        \bottomrule 
    \end{tabular*}
\end{table*}

\begin{table}[t]
    \centering
    \caption{Evaluation results of visual grounding in $Acc@0.5$ on RefCOCO, RefCOCO+, and RefCOCOg datasets.}
    \label{strongtool}
    \scalebox{0.8}{%
        \begin{tabular}{@{}lccccc@{}}
            \toprule
            \multirow{2}{*}{Method} 
            & \multicolumn{2}{c}{RefCOCO} 
            & \multicolumn{2}{c}{RefCOCO+} 
            & \multicolumn{1}{c}{RefCOCOg} \\
            \cmidrule(lr){2-3} \cmidrule(lr){4-5} \cmidrule(lr){6-6}
            & testA & testB & testA & testB & test \\
            \midrule 
            Qwen2.5-VL-7B \cite{qwen2.5VL} & 92.5 & 85.4 & 89.1 & 76.9 & 87.2 \\
            Qwen2.5-VL-72B \cite{qwen2.5VL} & {94.6} & 89.7 & 92.2 & \textbf{83.7} & 90.3 \\
            \hline
            EVF-SAM (tool) \cite{evf-sam} & 94.2 & {90.3} & 90.2 & 81.7 & 88.9 \\
            VG-Refiner (Ours) & \underline{95.0} & \underline{90.7} & \underline{92.7} & {83.0} & \underline{90.6} \\
            \rowcolor{gray!10}
            $\Delta$ vs. EVF-SAM (tool) 
            & +0.8 & +0.4 & +2.5 & +1.3 & +1.7 \\
            \rowcolor{gray!10}
            $\Delta$ vs. Qwen2.5-VL-7B 
            & +2.5 & +5.3 & +3.6 & +6.1 & +3.4 \\
            \hline

            Rex-omni (tool) \cite{rex_omni} & 89.4 & 83.6 & 85.1 & 72.9 & 83.9 \\
            VG-Refiner (Ours) & 92.6 & 85.8 & 89.9 & 76.6 & 88.2 \\
            \rowcolor{gray!10}
            $\Delta$ vs. Rex-omni
            & +3.2 & +2.2 & +4.8 & +3.7 & +4.3 \\
            \rowcolor{gray!10}
            $\Delta$ vs. Qwen2.5-VL-7B
            & +0.1 & +0.4 & +0.8 & -0.3 & +1.0 \\

            \hline

            UNINEXT-H (tool) \cite{uninext} & 94.3 & 91.5 & 89.6 & 79.9 & 89.2 \\
            VG-Refiner (Ours) & \textbf{95.6} & \textbf{92.3} & \textbf{92.8} & \underline{83.2} & \textbf{91.4} \\
            \rowcolor{gray!10}
            $\Delta$ vs. UNINEXT
            & +1.3 & +0.8 & +3.2 & +3.3 & +2.2 \\
            \rowcolor{gray!10}
            $\Delta$ vs. Qwen2.5-VL-7B
            & +3.1 & +6.9 & +3.7 & +6.3 & +4.2 \\

            \bottomrule 
        \end{tabular}
    }
\end{table}

\begin{table*}[t]
    \centering
    \caption{Performance comparison of different models under weak and strong tool conditions across RefCOCO, RefCOCO+, and RefCOCOg benchmarks under PiTER .}
    \label{tab:good_bad_tools}
    \scalebox{0.76}{
    \begin{tabular}{lccccccccccccccc}
        \toprule
        \multirow{2}{*}{Methods with PiTER} 
        & \multicolumn{3}{c}{RefCOCO testA (\%)} 
        & \multicolumn{3}{c}{RefCOCO testB (\%)} 
        & \multicolumn{3}{c}{RefCOCO+ testA (\%)} 
        & \multicolumn{3}{c}{RefCOCO+ testB (\%)} 
        & \multicolumn{3}{c}{RefCOCOg test (\%)} \\
        \cmidrule(lr){2-4} \cmidrule(lr){5-7} \cmidrule(lr){8-10} \cmidrule(lr){11-13} \cmidrule(lr){14-16}
         & ${Acc}$ & ${NSRI_w}$ & ${CCR}$ 
         & ${Acc}$ & ${NSRI_w}$ & ${CCR}$ 
         & ${Acc}$ & ${NSRI_w}$ & ${CCR}$ 
         & ${Acc}$ & ${NSRI_w}$ & ${CCR}$ 
         & ${Acc}$ & ${NSRI_w}$ & ${CCR}$ \\
        \midrule
        \multicolumn{16}{c}{\textbf{Weak Tool Conditions}} \\
        Grounding DINO T \cite{grounding-dino} & 49.9 & -- & -- & 37.8 & -- & -- & 50.0 & -- & -- & 38.7 & -- & -- & 54.6 & -- & -- \\
        \midrule
        Qwen2.5-VL-7B & 90.8 & 71.7 & 82.9 & 81.9 & 62.0 & 72.4 & 86.1 & 63.9 & 75.0 & 70.3 & 45.7 & 54.7 & 82.7 & 56.3 & 66.7 \\
        Qwen2.5-VL-32B & 89.9 & 69.4 & 80.3 & 82.1 & 61.1 & 71.7 & 84.2 & 59.2 & 69.0 & 71.3 & 45.1 & 54.0 & 84.1 & 54.5 & 65.9 \\
        
        REVPT & 71.2 & 36.9 & 46.7 & 61.4 & 31.4 & 41.5 & 64.1 & 24.4 & 32.0 & 53.2 & 19.5 & 27.9 & 67.3 & 23.4 & 32.1 \\

        VG-Refiner & \textbf{92.9} & \textbf{75.0} &\textbf{ 87.2} & \textbf{85.6} & \textbf{67.0} & \textbf{78.4} & \textbf{89.2 }& \textbf{68.9} & \textbf{80.7} & \textbf{75.8} & \textbf{51.7} & \textbf{62.6} & \textbf{86.6} & \textbf{61.9} & \textbf{73.5} \\
        \midrule
        \multicolumn{16}{c}{\textbf{Strong Tool Conditions}} \\
         EVF-SAM & 94.2 & -- & -- & 90.3 & -- & -- & 90.2 & -- & -- & 81.7 & -- & -- & 88.9 & -- & -- \\
        \midrule
        Qwen2.5-VL-7B & 92.7 & 22.2 & 39.9 & 86.3 & 10.6 & 27.7 & 89.8 & 30.1 & 48.4 & 77.4 & 10.9 & 27.9 & 86.1 & 19.5 & 33.9 \\
        Qwen2.5-VL-32B & \textbf{95.1} & 14.7 & 24.9 & \textbf{90.8} & 7.6 & 17.5 & \underline{92.4} & 19.8 & 33.3 & \textbf{83.6} & 8.9 & 20.7 & \textbf{90.6} & 14.7 & 23.6 \\
        REVPT & 88.9 & -3.6 & 7.5 & 82.9 & -7.8 & 3.2 & 87.2 & 2.0 & 10.5 & 77.1 & -1.6 & 8.0 & 86.4 & 0.9 & 9.0 \\

        VG-Refiner & \underline{94.8} & \textbf{30.6} & \textbf{47.8} & \underline{89.9} & \textbf{17.5} & \textbf{30.5} & \textbf{92.5} & \textbf{36.4} & \textbf{54.5} & \underline{81.9} & \textbf{16.3} & \textbf{32.4} & \underline{89.3} & \textbf{25.7} & \textbf{39.4 }\\
        \bottomrule
    \end{tabular}
    }
\end{table*}

\begin{table*}[t]
    \centering
    \small 
    \caption{Comparison of reasoning grounding performance. The tools used in PiTER and VG-Refiner are both EVF-SAM.}
    \label{tab:t3}
    \scalebox{1}{
    \begin{tabular*}{\linewidth}{@{\extracolsep{\fill}}lccccc@{}}
        \toprule
        Method & VLM-R1-3B & Qwen2.5-VL-7B & Qwen2.5-VL-7B (PiTER) & EVF-SAM (tool) & \textbf{VG-Refiner (Ours)} \\ 
        \midrule
        LISA-Test & 63.1 & 67.1 & 65.7 & 48.9 & \textbf{68.5} \\
        \bottomrule
    \end{tabular*}
    }
\end{table*}

\subsection{Agentic RL Training}
\textbf{Group Relative Policy Optimization (GRPO).} We enable our VG-Refiner to evolve via agentic RL, where the model actively invokes external tool results during the RL process. We do not rely on the cold start data for first supervised finetuning, as the baseline model already demonstrates strong refinement ability (Table~\ref{tab:good_bad_tools}). To realize this training scheme, we adopt GRPO~\cite{deepseek-r1} on the pretrained model. During each update, the policy model performs multiple rollouts to generate a set of $G$ reasoning trajectories. Each trajectory $o_i$ in the TrRGR paradigm is represented as $\{T_i, A_i, F_i, R_i\}$. We then evaluate the generated trajectories using our rule-based reward function, which assigns a scalar reward to every $o_i \in \{o_1, o_2, \ldots, o_G\}$. To further enhance the model’s refinement capability, we incorporate two complementary types of rewards described below.

\textbf{Format Reward.} To encourage the model to produce structured and interpretable reasoning traces, we introduce a format reward that verifies whether the generated output follows the required reasoning schema. Specifically, the required output is ``
\textless \textit{think}\textgreater~$P^1_\text{CoT}$ \textless \textit{/think}\textgreater,
\textless \textit{rethink}\textgreater~$P^2_\text{CoT}$ \textless \textit{/rethink}\textgreater,
and \textless \textit{answer}\textgreater~$\{$``bbox\_2d": [$x$\textsubscript{1}, $y$\textsubscript{1}, $x$\textsubscript{2}, $y$\textsubscript{2}]$\}$ \textless \textit{/answer}\textgreater.''
A positive reward of 1 is assigned for $R_{format}$ if all tags are correctly formatted; otherwise, the reward is set to zero.
This encourages the model to generate explicit refinement reasoning in a structured manner.

\textbf{Refinement Reward.} Since tool feedback may be reliable or erroneous, the reward must reflect these differing levels of difficulty. Notably, correcting an incorrect tool output is more challenging than confirming a correct one. To guide the model in both recognizing trustworthy predictions and refining unreliable ones, we introduce two components: a \textit{tool-confirmation reward} $R^G_{refine}$ and a \textit{tool-refinement reward} $R^W_{refine}$. Combined, they form our {refinement reward}, which enables the model to adaptively utilize tool feedback based on its quality. Let $\mathrm{IoU}_t$ and $\mathrm{IoU}_f$ denote the IoU scores of the {tool prediction} and the {final model prediction} with the GT, respectively, shown in Figure \ref{framework}. 
\begin{equation}
\begin{aligned}
R_{\text{refine}}^G &= 
\begin{cases}
0.5, & \mathrm{IoU}_t \ge 0.5 \ \text{and} \ \mathrm{IoU}_f \ge 0.5, \\[3pt]
0, & \text{otherwise},
\end{cases}
\\[8pt]
R_{\text{refine}}^W &= 
\begin{cases}
1, & \mathrm{IoU}_t < 0.5 \ \text{and} \ \mathrm{IoU}_f \ge 0.5, \\[3pt]
0, & \text{otherwise}.
\end{cases}
\end{aligned}
\end{equation}
We adopt this coarse piecewise reward to ensure stability and avoid reward hacking. A fixed 0.5 reward prevents overfitting to noisy IoU fluctuations when the tool is already correct, while 1 reward is given only for correcting wrong tool outputs, which represents a truly substantive refinement.

\subsection{Refinement Ability Evaluation}
\textbf{PiTER Evaluation Protocol.} We propose the PiTER evaluation protocol to systematically assess the refinement capability of different models. To ensure a fair comparison, we design a unified system prompt shown in Figure \ref{user_prompt}, Prompt-injected Tool for Enhancement and Refinement (PiTER), which directly injects the tool results as external references in the input stage. All models are required to perform inference in a single stage, regardless of their original paradigms or how they integrate tool feedback during reasoning. This removal of all intermediate verbal reasoning steps can reveal the intrinsic refinement capability of each model. Additionally, we also require them to generate predictions in a JSON format consistent with the Qwen2.5-VL pretraining stage to accomplish the grounding tasks. 

\textbf{Evaluation Metrics.} We propose two complementary metrics to comprehensively evaluate the refinement performance of the TrRGR model: critical correct rate (CCR) and normalized signed relative IoU (NSRI). The former metric measures the proportion of successful refinements made by the model on tool-failed cases. In contrast, the latter evaluates the IoU improvement relative to the maximum possible gain, reflecting how effectively the model utilizes the available refinement space. For each sample $i$, we define the total tool wrong cases as $S_{{w}} = \{ i \mid \text{IoU}^{{i}}_t < 0.5 \}$ and the CCR can be calculated shown below:
\begin{equation}
\text{CCR} = 
\frac{
\sum_{i \in S_{{w}}} 
\mathbb{I}[\text{IoU}^{{i}}_f \ge 0.5]
}{
|S_{{w}}|
},
\end{equation}
where $\mathbb{I}[\cdot]$ is the indicator function. In parallel, we define an NSRI change $g_i$ as:
\begin{equation}
g_i = 
\begin{cases}
\dfrac{\text{IoU}^i_f - \text{IoU}^{{i}}_t}{1 - \text{IoU}^{{i}}_t}, 
& \text{if } \text{IoU}^i_f > \text{IoU}^{{i}}_t   \\[8pt]
\dfrac{\text{IoU}^i_f - \text{IoU}^{{i}}_t}{\text{IoU}^{{i}}_t}, 
& \text{if } \text{IoU}^i_f < \text{IoU}^{{i}}_t  \\[8pt]
0, & \text{otherwise.}
\end{cases}
\end{equation}
This normalization ensures that $g_i \in [-1, 1]$, where a positive value represents the possible improvement, vice versa. To focus on the model’s behavior when the tool fails, we compute the average value of $g_i$ over the subset $S_{{w}}$ as a final metric, defined as $NSRI_w$.

%% file: sec/4_experiment.tex
\section{Experiment}
\subsection{Experimental Settings}


\textbf{Datasets.} Following the VLM-R1 \cite{vlm-r1}, to test the in-domain performances, we evaluate our VG-Refiner on the RefCOCO/+/g datasets and test the out-of-domain performances on LISA-grounding \cite{lisa} test sets, which are two classic referring reasoning grounding datasets. To further demonstrate that our model has the general QA capabilities, we benchmark it on MMbench ~\cite{mmbench}, OCRBench ~\cite{ocrbench}, RealWordQA ~\cite{realwordqa}, ChartQA ~\cite{chartqa}, and MMStar  ~\cite{mmstar}.

 
\begin{figure*}[ht]
	\centering	\includegraphics[width=\textwidth]{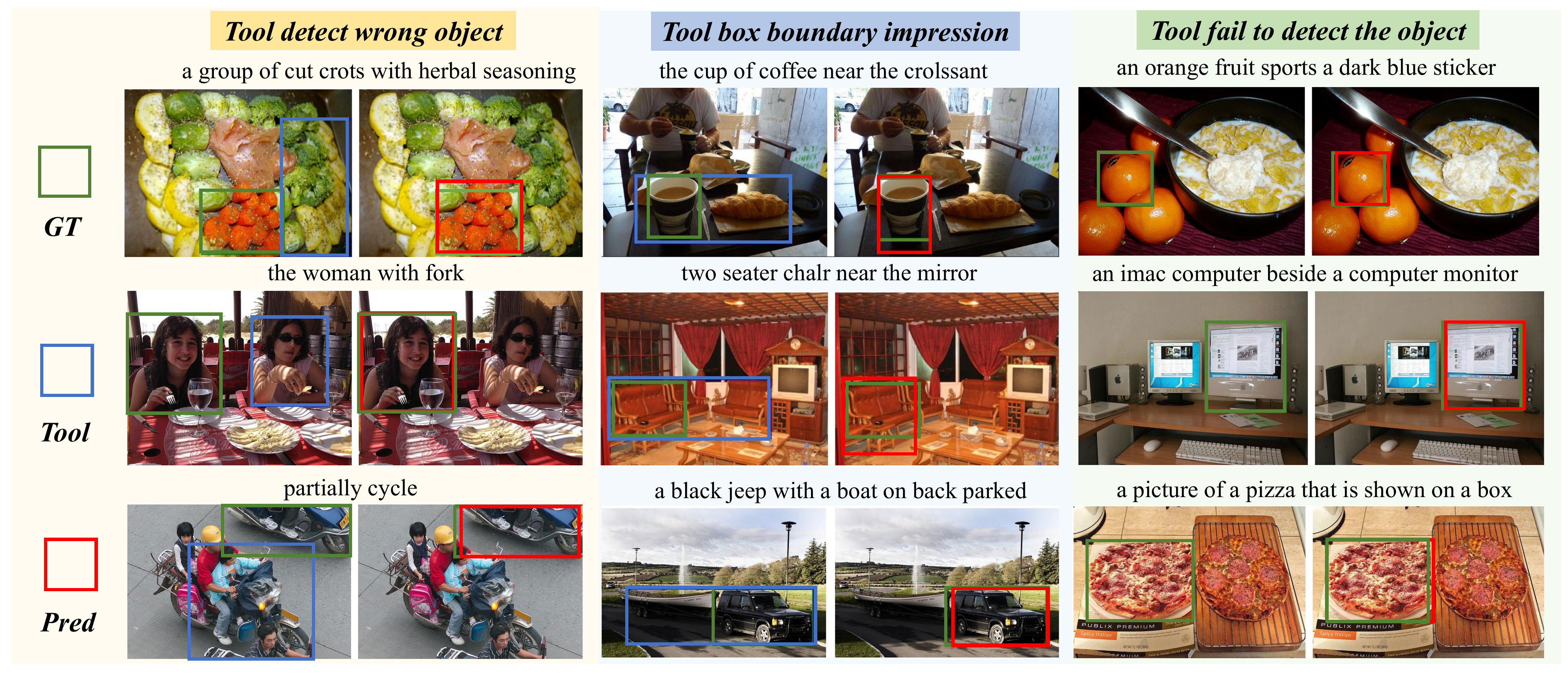}
	\caption{Visualization of VG-Refiner handling three representative types of tool-induced errors in TrRGR.  The first two grounding error categories often occur in a good tool, EVF-SAM \cite{evf-sam}, whereas the third occurs in the not fine-tuned Grounding DINO T \cite{grounding-dino}. }
	\label{vis1}
\end{figure*}

\textbf{Implementation Details.} Following REVPT~\cite{revpt}, we adopt Qwen2.5-VL-7B as our baseline model. Our training framework is built upon the open-source verl~\cite{verl} and vLLM~\cite{vllm} implementations. Following ~\cite{seg-zero}, we select a total of 9K samples from the RefCOCOg dataset for training, where the tool outputs are partially derived from a strong expert model, EVF-SAM~\cite{evf-sam}, and partially from a weak tool, no fine-tuned Grounding DINO T (GD-T)~\cite{grounding-dino}. Although EVF-SAM is designed for referring comprehension segmentation (RES), its generated bounding boxes converted from masks are sometimes imprecise, being either too tight or too loose around the target due to mask quality artifacts. This property makes it particularly suitable for our refinement setting, as the model can further adjust such imperfect yet correct predictions to achieve higher IoU scores. For the training configuration, we perform 8 rollouts per sample, with a training batch size of 128 and a rollout batch size of 32. We set the learning rate to $1.0\times10^{-6}$ and use a sampling temperature of 1.0. The model is trained for one epoch to preserve its general QA capabilities and can be applied out of domain. All the experiments are conducted on 4 NVIDIA A100 GPUs.

\textbf{Metrics.} We adopt the standard $Acc@0.5$ metric, measuring the proportion of predictions with IoU above 0.5 across all the test samples. To assess refinement ability, we use $NSRI_w$ and $CCR$, focusing on samples where the tool outputs are incorrect (i.e., $\text{IoU}<0.5$). All models are evaluated with official checkpoints under a unified evaluation protocol and codebase for fairness.

\subsection{Main Results}
Table~\ref{tabt1} presents the referring grounding results on the RefCOCO series benchmarks. During inference, VG-Refiner calls the EVF-SAM model as an expert tool and explicitly analyzes its outputs in the rethink stage. By integrating tool feedback with its own reasoning process, VG-Refiner achieves SOTA performance, comparable even to 72B-scale models. With the assistance of the tool, our model exhibits substantial performance gains over the original Qwen2.5-VL-7B baseline, consistently outperforming the tool's outputs across all benchmarks. These results demonstrate that VG-Refiner is capable of discerning when to follow or refine the tool predictions, rather than blindly trusting expert tool outputs. This TrRGR paradigm surpasses traditional verbal reasoning and SFT approaches, achieving a co-evolution between the tool and the refiner.

\begin{table}[t]
    \centering
    \caption{General QA evaluation across challenging benchmarks.}
    \scalebox{0.8}{
    \begin{tabular}{lccc}
        \toprule
        Dataset & Qwen2.5-VL-7B & REVPT & VG-Refiner \\
        \midrule
        MMBench-EN\textsubscript{DEV} & {80.0} & 50.7 & 79.0 \\
        MMBench-CN\textsubscript{DEV} & {82.2} & 53.9 & {82.5} \\
        MMBench-EN\textsubscript{DEV-V1.1} & {80.0} & 52.5 & 79.1 \\
        MMBench-CN\textsubscript{DEV-V1.1} & {81.4} & 55.1 & {81.5} \\
        OCRBench & {886} & 880 & {886} \\
        Chart QA & {86.2} & 84.0 & {86.2} \\
        Realworld QA & {68.1} & 51.4 & 67.5 \\
        MMStar  & {60.8} & 45.8 & 60.1 \\
        \bottomrule
    \end{tabular}
    }
    \label{tab:t4}
\end{table}

As shwon in Table \ref{strongtool}, to further evaluate the robustness of VG-Refiner under different strong tool conditions, we additionally test it using tool predictions from Rex-Omni~\cite{rex_omni} and UNINEXT-H~\cite{uninext}, both of which achieve strong performance on the REC task. Despite being trained solely on EVF-SAM tool outputs, our model demonstrates strong scalability and delivers consistent improvements across all five test sets. Notably, when provided with high-quality tool results such as those from UNINEXT-H, it can even surpass the performance of models at the 72B scale.

\begin{table}[t]
    \centering
    \caption{Ablation study on reward design under PiTER protocol using EVF-SAM tool result.}
    \label{tab:t5}
    \scalebox{0.74}{
    \begin{tabular}{lcccc}
        \toprule
        \multirow{2}{*}{Dataset} 
        & \multicolumn{2}{c}{IoU Reward} 
        & \multicolumn{2}{c}{Refinement Reward} \\
        \cmidrule(lr){2-3} \cmidrule(lr){4-5}
        & $Acc$ (\%) & $NSRI_w$ (\%) 
        & $Acc$ (\%) & $NSRI_w$ (\%)  \\
        \midrule
        RefCOCO testA   & 93.9 & 24.4 & 94.8 & 30.6 \\
        RefCOCO testB   & 88.8 & 16.6 & 89.9 & 17.5 \\
        RefCOCO+ testA  & 91.4 & 32.2 & 92.4 & 36.4 \\
        RefCOCO+ testB  & 80.1 & 15.2 & 81.9 & 16.3 \\
        RefCOCOg test   & 87.2 & 22.0 & 89.2 & 25.7 \\
        \bottomrule
    \end{tabular}
    }
\end{table}

\begin{figure*}[t]
	\centering	\includegraphics[width=\textwidth]{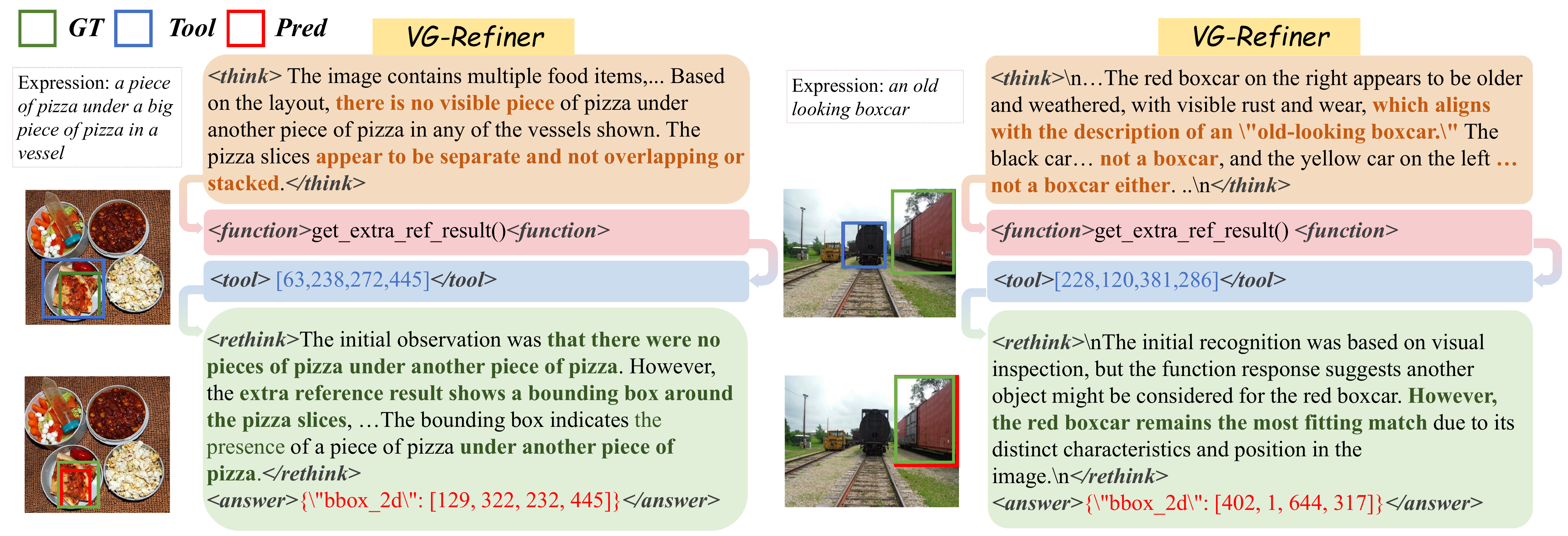}
	\caption{Visualization of the overall reasoning paradigm, first performing self-thinking and then re-thinking based on the tool outputs. }
	\label{vis2}
\end{figure*} 
Table~\ref{tab:good_bad_tools} reports the refinement capabilities under the PiTER protocol across five test sets of the RefCOCO series benchmarks. Under weak tool conditions (i.e., use not finetuned GD-T tool), VG-Refiner significantly outperforms both 32B- and 7B-sacle models, demonstrating superior robustness to unreliable tool predictions. In contrast, REVPT shows limited refinement ability due to the absence of explicit error-handling mechanisms. Notably, our model can effectively refine weak tool outputs to reach performance comparable to its own grounding ability. Under strong tool conditions (i.e., use EVF-SAM), REVPT fails to preserve the tool’s original performance and even degrades results, as reflected by a negative $NSRI_w$, indicating that it tends to optimize tool outputs in a harmful direction. This further highlights VG-Refiner’s advantage in adaptively handling both reliable and unreliable tool feedback.

We further report the out-of-domain performances on the LISA-Grounding test set, as shown in Table \ref{tab:t3}, where the referring phrases require stronger reasoning capabilities. In such cases, EVF-SAM serves as a weak tool since it has not been fine-tuned on this dataset. We evaluate all models in a zero-shot setting. Our VG-Refiner still surpasses the original Qwen2.5-VL-7B baseline, demonstrating its ability to identify and refine incorrect tool outputs, even under unseen and reasoning-intensive conditions.

\begin{table}[t]
    \centering
    \caption{Ablation study on Think-Rethink mechanism in $Acc$ (\%).}
    \label{tab:t6}
    \scalebox{0.83}{
    \begin{tabular}{lcccc}
        \toprule
        \multirow{2}{*}{Method} & \multicolumn{1}{c}{RefCOCO} & \multicolumn{2}{c}{RefCOCO+} & \multicolumn{1}{c}{RefCOCOg} \\
        \cmidrule(lr){2-2} \cmidrule(lr){3-4} \cmidrule(lr){5-5}
         & testB & testA & testB & test \\
        \midrule
        1\textsubscript{st} stage tool prompt & 86.3 & 89.8 & 77.4 & 86.1 \\
        2\textsubscript{nd} stage tool prompt & 89.8 & 90.9 & 81.9 & 88.9 \\
        think  & 88.7 & 90.2 & 78.6 & 88.7 \\
        think-rethink & \textbf{90.7} & \textbf{92.7} & \textbf{83.0} & \textbf{90.5} \\
        \bottomrule
    \end{tabular}
    }
\end{table}

\subsection{Quality Assessment}
\textbf{Tool Refinement.} Figure~\ref{vis1} presents qualitative visualizations of our {VG-Refiner} under various tool feedback conditions. For incorrect tool predictions, our model can rethink the tool’s feedback and perform self-analysis to produce more accurate and robust grounding results.

\textbf{Rethink Analysis.} Figure~\ref{vis2} illustrates that the \textit{rethink} CoT process serves in refining and achieving the final correct predictions, highlighting the key advantage of the TrRGR paradigm. For instance, in the boundary ambiguity case, although VG-Refiner initially fails to detect the \textit{“pizza under a big piece of pizza in a vessel”} during the think stage, the additional reference provided by the tool introduces contextual visual cues, enabling it to localize the true referred object accurately. This observation aligns with recent work TreeVGR~\cite{traceable}, which enhances visual grounded reasoning by generating an evidence box to provide an additional spatial focus. Secondly, for completely incorrect tool predictions, VG-Refiner can reject the unreliable feedback and identify the true target during the rethinking process, demonstrating strong robustness against disturbances from erroneous tool outputs. 



\subsection{Ablation Study}
\textbf{General Visual QA.} Since we only use a small amount of training data to enhance the refinement capability of VG-Refiner, we further investigate its original general ability. As shown in Table \ref{tab:t4}, we compare the visual QA performance of our model with the baseline Qwen2.5-VL-7B and the TiVR model REVPT. Our VG-Refiner preserves the general visual QA capability and even achieves slight performance gains on some benchmarks.

\begin{table}[t]
    \centering
    \caption{Evaluation on using Qwen2.5-VL-7B as a tool in PiTER protocol on RefCOCOg test subset. The 84.2 accuracy is our reproduced results of Qwen2.5-VL-7B using JSON format output.}
    \label{tab:refcocog_ablation}
    \scalebox{0.85}{
    \begin{tabular}{lcccc}
        \toprule
        Method in PiTER & Acc (\%) & $NSRI_w$ (\%) & $CCR$ (\%) \\
        \midrule
        Qwen2.5-VL-32B & \textbf{86.0} vs 84.2 & 11.0 & 14.9 \\
        Qwen2.5-VL-7B  & 83.7 vs 84.2 & 8.2 & 11.4 \\
        VG-Refiner & 85.7 vs 84.2 & \textbf{12.6} & \textbf{16.0} \\
        \bottomrule
    \end{tabular}
    }
\end{table}
\textbf{Reward Design.} The design of the refinement reward plays a crucial role in enhancing the correction capability of VG-Refiner. As shown in Table \ref{tab:t5}, both reward types are verifiable rewards computed by comparing the prediction with the GT. However, unlike the direct IoU reward, the proposed refinement reward explicitly considers the circumstances of tool feedback. When the tool prediction is incorrect but the model’s final prediction is correct, the model receives the full reward. Conversely, if the tool prediction is correct but the model only marginally refines it, the reward is set to 0.5. This hierarchical reward mechanism encourages the model to focus on correcting inaccurate tool outputs, leading to higher overall accuracy and NSRI\textsubscript{w}. The results confirm the effectiveness of this design, showing consistent performance gains across all benchmarks.

\textbf{Think-rethink Paradigm.} As shown in the first and second rows of Table~\ref{tab:t6}, we first investigate the effect of tool feedback positioning using the baseline model {Qwen2.5-VL-7B}. The first row presents the evaluation under the {PiTER} setting, while the second row corresponds to a two-stage conversation setup. In both conversations, the model is required to output the bounding box in JSON format, but the tool results are introduced only in the second stage as external feedback. These results demonstrate that incorporating the second-stage tool prompt yields a substantial performance improvement, as the model can jointly exploit its initial recognition and the tool’s prediction to achieve more accurate grounding. Furthermore, under two rounds of conversations, the third and fourth rows highlight the significance of the {rethink stage}, which allows the model to further refine the tool outputs, revealing stronger self-correction and reasoning capabilities.

\textbf{Self Improvement.} Our {VG-Refiner} is capable of refining arbitrary tool outputs. In this experiment shown in Table \ref{tab:refcocog_ablation}, we use the reproduced baseline results of {Qwen2.5-VL-7B} as the tool feedback under the PiTER protocols. While the original Qwen2.5-VL-7B fails to improve when provided with its own predictions as prompts, our VG-Refiner successfully identifies and corrects these suboptimal results, leading to consistent performance gains.


%% file: sec/5_conclusion.tex
\section{Conclusion}
In this paper, we propose a tool-refined referring grounded reasoning paradigm to address the hallucination and negative optimization issues caused by unreliable or erroneous tool feedback in tool-integrated reasoning for visual grounding tasks. We introduce VG-Refiner, an agentic reinforcement learning model trained with a limited amount of data. By perceiving general visual question answering capabilities, VG-Refiner enhances its refinement ability toward external tool feedback through a carefully designed refinement reward and a two-stage think–rethink framework. Our model can serve as a foundation for future multi-round tool-calling frameworks, enabling more robust reasoning and fine-grained comprehension of small objects.

%% file: sec/X_suppl.tex
\clearpage
\setcounter{page}{1}
\maketitlesupplementary

\section{The Choice of Tools}
\label{sec:rationale}
\begin{figure}[t]
	\centering	\includegraphics[width=0.46\textwidth]{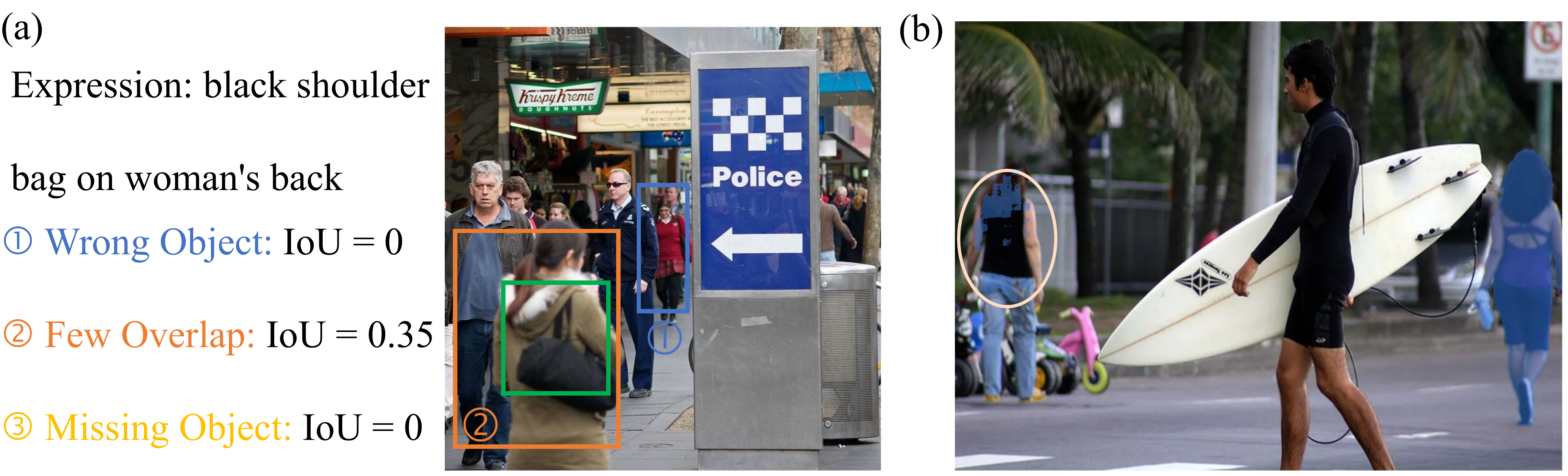}
	\caption{(a)The common three errors in the referring expression comprehension task. (b) The output mask of the RES model with expression: A lady is walking away from the surfer.}
	\label{error_category}
\end{figure}

\textbf{Tool Error Category.} As shown in Figure \ref{error_category}, the referring expert tool may fail in three ways: 
(1) \textit{wrong object localization}, 
(2) \textit{boundary imprecision} where the predicted mask has only minor overlap with the target, and 
(3) \textit{missing object localization}. 
For objects that are indeed present in the image, the weak tool often outputs a null box because it is not fine-tuned to understand complex referring expressions. In contrast, even a strong tool can occasionally localize an incorrect object, as no tool is perfect.
\begin{table}[t]
    \centering
    \caption{Evaluation results of visual grounding in $ACC@0.5$ on RefCOCO, RefCOCO+, and RefCOCOg datasets.}
    \label{tab:t12}
    \scalebox{0.8}{%
        \begin{tabular}{@{}lccccc@{}}
            \toprule
            \multirow{2}{*}{Method} 
            & \multicolumn{2}{c}{RefCOCO} 
            & \multicolumn{2}{c}{RefCOCO+} 
            & \multicolumn{1}{c}{RefCOCOg} \\
            \cmidrule(lr){2-3} \cmidrule(lr){4-5} \cmidrule(lr){6-6}
            & testA & testB & testA & testB & test \\
            \midrule 
            Grounding DINO T \cite{grounding-dino} & 49.9 & 37.8 & 50.0 & 38.7 & 54.6 \\
            Qwen2.5-VL-7B \cite{qwen2.5VL} & 92.5 & 85.4 & 89.1 & 76.9 & 87.2 \\
            REVPT \cite{revpt}+LLMDET \cite{llmdet} & 77.3 & 68.5 & 71.1 & 59.4 & 67.2 \\
            REVPT+EVFSAM \cite{evf-sam} & 90.1 & 75.1 & 75.5 & 66.6 & 75.3 \\
            VG-Refiner+EVF-SAM & 95.0 & 90.7 & 92.7 & 83.0 & 90.6 \\
            \bottomrule 
        \end{tabular}
    }
\end{table}

\textbf{RES Model as a Strong Tool.}
We adopt EVF-SAM \cite{evf-sam} as the strong referring expression segmentation (RES) tool, given its state-of-the-art performance. Since RES outputs masks, we convert them into bounding boxes by taking the extreme left, top, right, and bottom pixel coordinates. However, in ambiguous scenarios with multiple similar objects, the RES model may produce masks spanning several regions or containing artifacts, shown in Figure~\ref{error_category}(b). Such cases lead to boundary imprecision when converting masks to boxes, thereby increasing the difficulty for the refiner model.

\textbf{Weak Tool.} We use the non-finetuned Grounding DINO-T as the weak tool, as it struggles with complex and long referring expressions while still performing reasonably well on simple category-name queries. When evaluated on the RefCOCO/+/g test sets, it achieves an average mean accuracy of only around 0.4, indicating that its predictions are unreliable and relatively random, as shown in the Table. Such behavior makes it well-suited to serve as the weak tool in our refiner framework.

\section{More Benchmark Results}
\textbf{REVPT Benchmark Results.} We benchmark the REVPT model~\cite{revpt} using two types of expert tools, following its multi-round tool-calling paradigm. The overall performance remains relatively low, primarily because the model often fails to generate outputs in the required format and the REC model inputs are not sufficiently specific when calling parameters. Moreover, REVPT’s predictions tend to closely track the tool outputs rather than refine them, which aligns with our earlier observations.

\textbf{More Results on Self-Improvement Experiments.} The prompt tool results used in the self-improvement experiment are generated by our reproduced Qwen2.5-VL-7B, which is required to output in the JSON format. We further evaluate on additional test sets to demonstrate that our VG-Refiner framework can be applied to any tool outputs, even those from the original baseline, while achieving performance comparable to a 32B-scale model.

\begin{table}[t]
    \centering
    \caption{Performance comparison of different models under self tool conditions on RefCOCO+ (testA and testB) benchmarks under PiTER.}
    \label{tab:self_tools_refcoco_plus}
    \scalebox{0.72}{
    \begin{tabular}{lcccccc}
        \toprule
        \multirow{2}{*}{Methods with PiTER} 
        & \multicolumn{3}{c}{RefCOCO+ testA (\%)} 
        & \multicolumn{3}{c}{RefCOCO+ testB (\%)} \\
        \cmidrule(lr){2-4} \cmidrule(lr){5-7}
         & ${Acc}$ & ${NSRI_w}$ & ${CCR}$ 
         & ${Acc}$ & ${NSRI_w}$ & ${CCR}$ \\
        \midrule
        \textbf{Self Tool Conditions} & 88.0 & -- & -- & 74.2 & -- & -- \\
        \midrule
        Qwen2.5-VL-7B & 88.0 & 7.4 & 10.8 & 73.5 & 6.8 & 8.6 \\
        Qwen2.5-VL-32B & 88.8 & 10.4 & 13.5 & \textbf{75.8} & 7.3 & \textbf{10.7} \\
        VG-Refiner-7B & \textbf{89.0} & \textbf{11.1} & \textbf{15.0} & {75.6} & \textbf{8.5} & {10.5} \\
        \bottomrule
    \end{tabular}
    }
\end{table}

\textbf{Refiner Compliance and Stability Metrics.} To assess whether the refiner faithfully preserves correct tool predictions while avoiding harmful degradation, we introduce two metrics: the \emph{Follow Correct Rate} and the \emph{Worsen Rate}. All definitions rely on the IoU relationship between the tool prediction $\text{IoU}^i_t$ and the refiner output $\text{IoU}^i_f$.

We first define the tool-correct set as $S_{c} = \left\{\, i \mid \text{IoU}^i_t \ge 0.5 \,\right\}.$
Within this set, the refiner is considered to follow the tool when its IoU remains nearly unchanged, i.e., $
\mathcal{F} = \left\{i \mid \left|\,\text{IoU}^i_f - \text{IoU}^i_t\,\right| < \epsilon, \right\}$ forming the follow set $\mathcal{F}$.
In parallel, we define the global worsen set
$
\mathcal{W} = \left\{\, i \mid \text{IoU}^i_f < \text{IoU}^i_t \,\right\},$ which includes all samples in which the refiner degrades the tool output. The \emph{Follow Correct Rate}  (FCR) measures how reliably the refiner preserves correct tool predictions:
\[
\text{FCR} = \frac{|\mathcal{F}|}{|S_{c}|} \times 100\%.
\]
The \emph{Worsen Rate} (WR) quantifies the overall fraction of degradations across the entire dataset:
\[
\text{WR} = \frac{|\mathcal{W}|}{N} \times 100\%,
\]
where $N$ is the total number of samples.

\begin{table}[t]
    \centering
    \caption{Comparison of Refiner 7B and Qwen32B on multiple referring datasets.}
    \label{tab:fcr}
    \scalebox{0.65}{
    \begin{tabular}{lcccccc}
        \toprule
        \multirow{2}{*}{Dataset} 
        & \multicolumn{3}{c}{VG-Refiner 7B} 
        & \multicolumn{3}{c}{Qwen2.5-VL-32B} \\
        \cmidrule(lr){2-4} \cmidrule(lr){5-7}
        & Acc (\%) & FCR (\%) & WR (\%) 
        & Acc (\%) & FCR (\%) & WR (\%) \\
        \midrule
        RefCOCO testA   & 95.0 & 96.7 & 2.2 & 95.1 & 96.7 & 2.5 \\
        RefCOCO testB   & 90.7 & 96.7 & 2.6 & 90.8 & 93.0 & 5.0 \\
        RefCOCOg test   & 90.5 & 95.7 & 2.9 & 90.6 & 91.4 & 5.7 \\
        RefCOCO+ testA  & 92.7 & 95.5 & 3.4 & 92.4 & 95.0 & 3.7 \\
        RefCOCO+ testB  & 83.0 & 94.3 & 4.8 & 83.6 & 90.4 & 7.2 \\
        \bottomrule
    \end{tabular}
    }
\end{table}

We report the FCR and WR results in Table~\ref{tab:fcr}. Compared with Qwen2.5-VL-32B under the PiTER protocol, our VG-Refiner more faithfully follows reliable tool outputs while avoiding further degradation of the tool predictions.

\begin{figure}[t]
    \centering
    \includegraphics[width=0.46\textwidth]{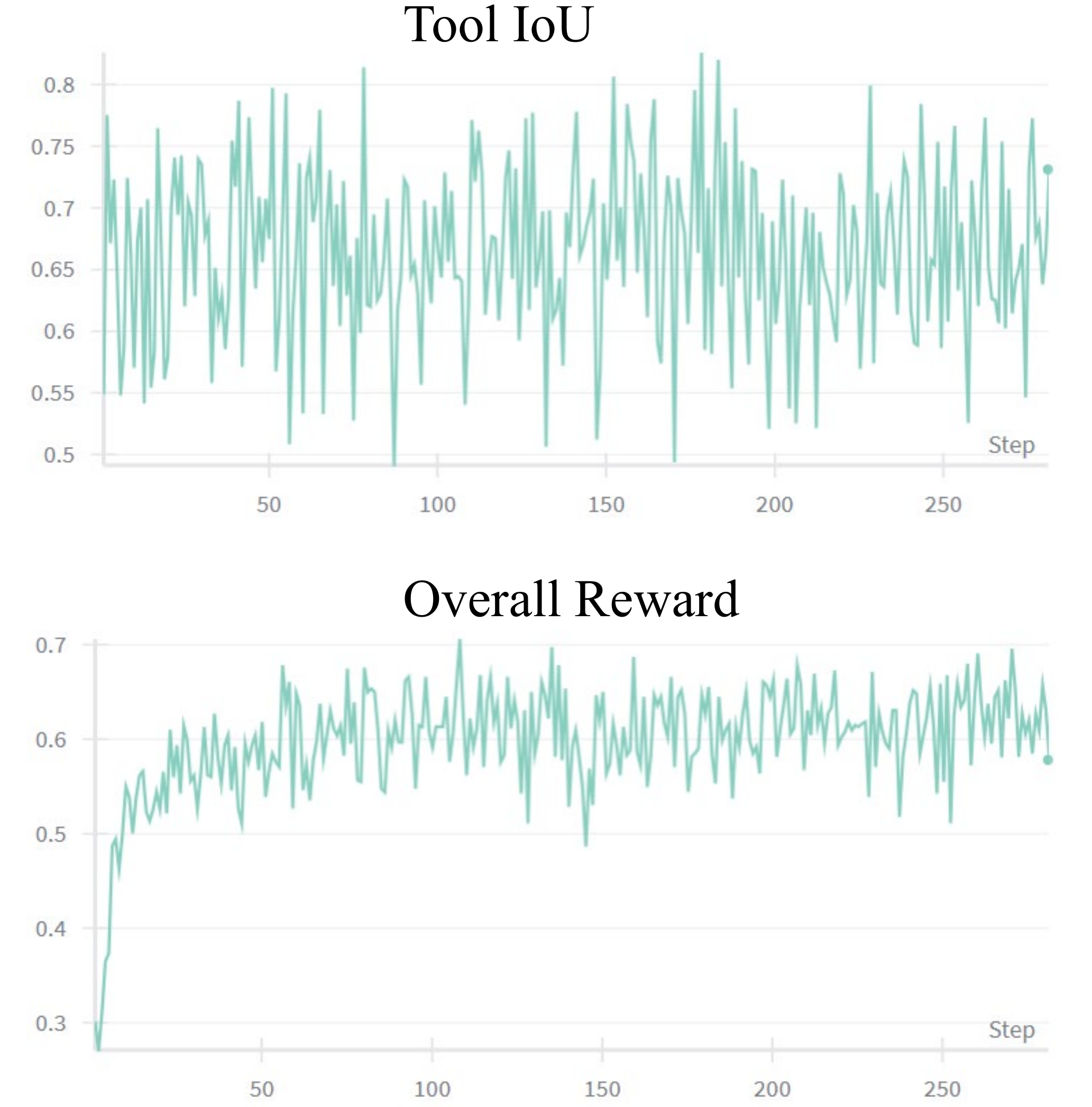}
    \includegraphics[width=0.46\textwidth]{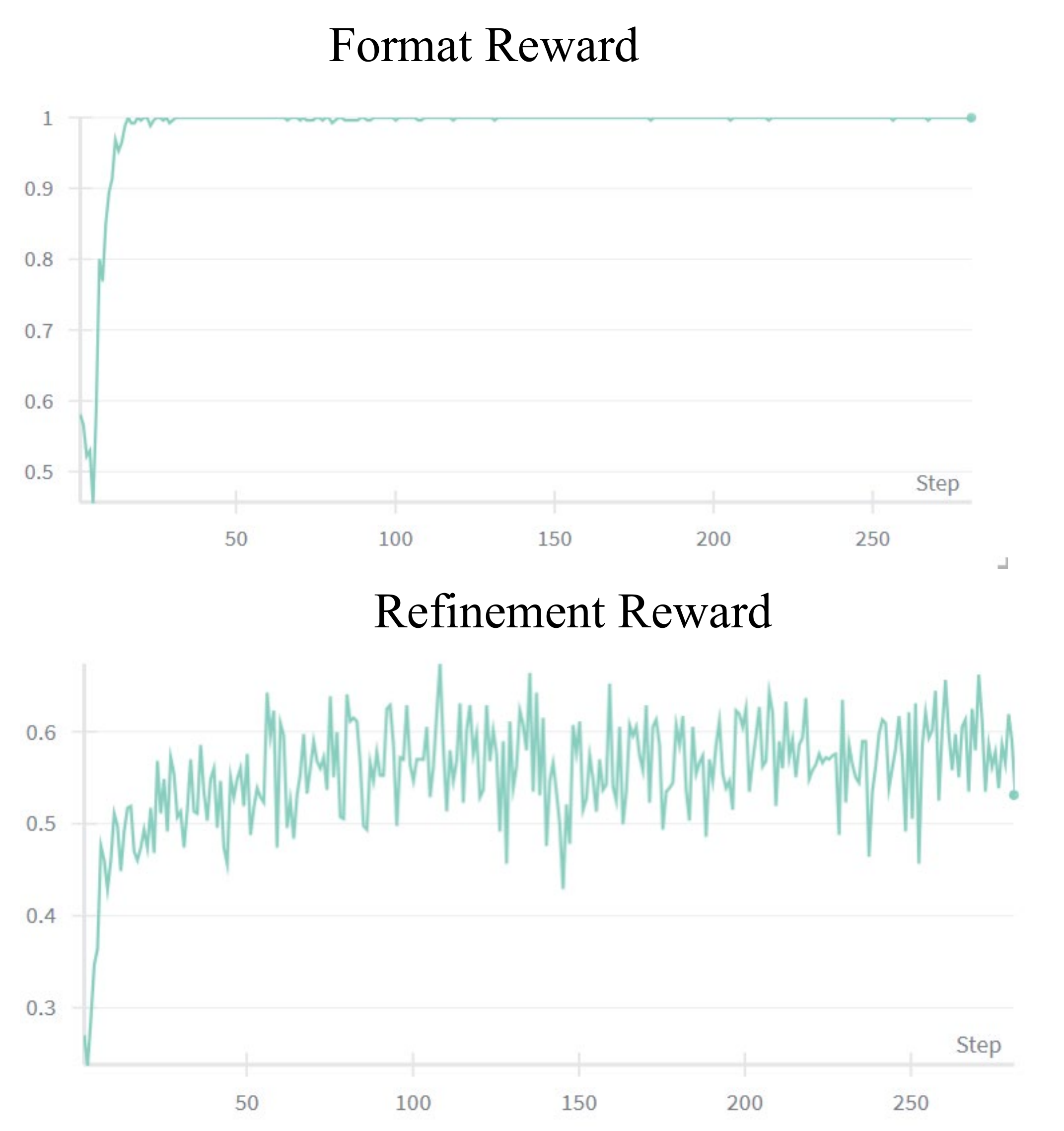}
    \caption{The reward changes with the training steps. We show the mean value across the batch.}
    \label{reward_curve}
\end{figure}

\begin{figure*}[t]
	\centering	\includegraphics[width=\textwidth]{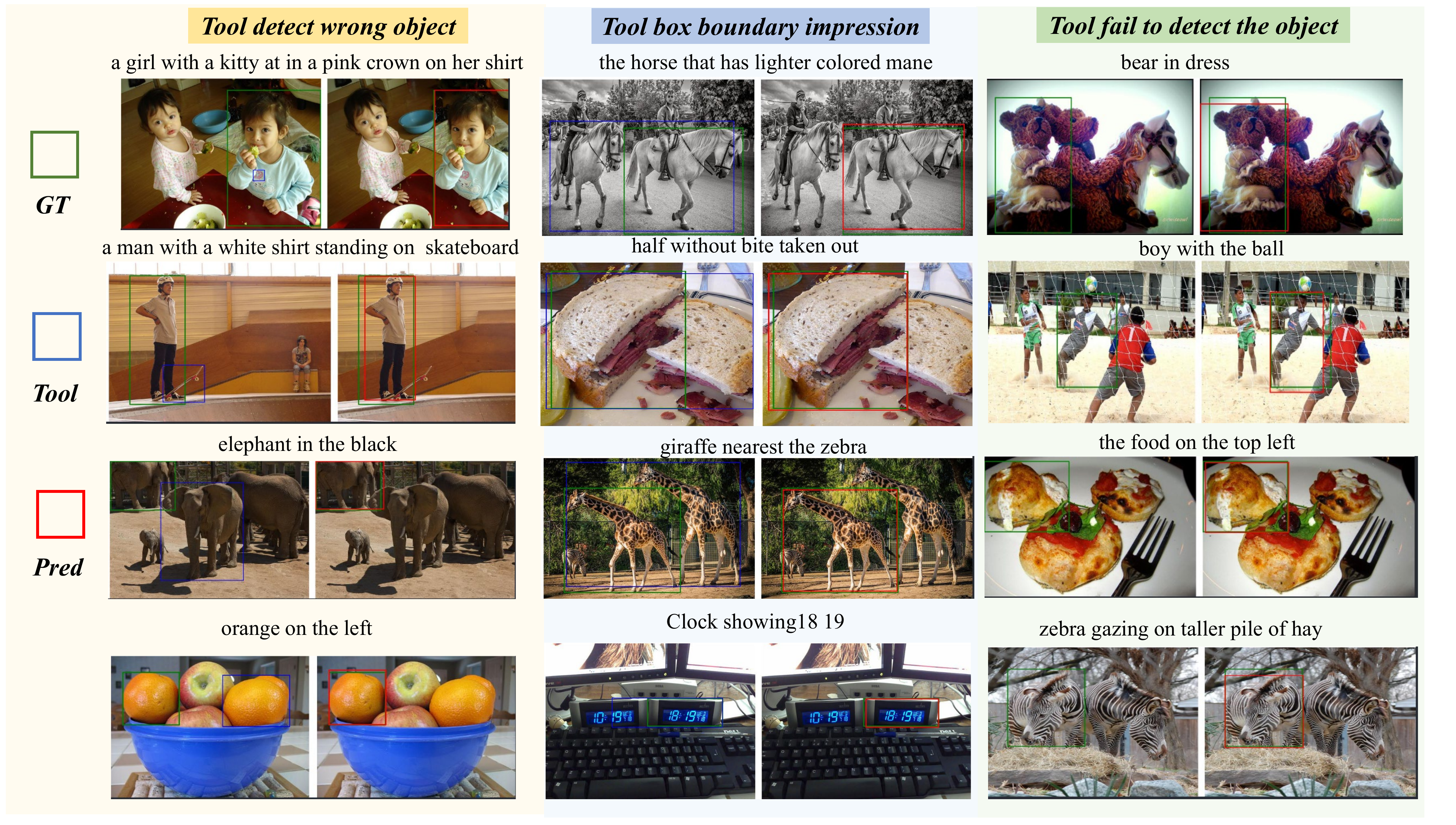}
	\caption{More visualization results.}
	\label{sup_vis_1}
\end{figure*} 
\section{Training Details}
\textbf{Reward Changes During Training }
We visualize the reward curves during training in Figure~\ref{reward_curve}. The refinement reward remains consistently above 0.5, indicating that the model effectively corrects incorrect tool outputs and receives the corresponding tool-refinement reward. The tool’s overall IoU hovers around 0.6, providing an appropriate level of difficulty and contributing to stable training dynamics. The format reward stays near 1 throughout training, demonstrating that the model reliably adheres to the required structured output format.

\textbf{Training Datasets.} We construct a 9k-sample training set by mixing inference results from EVF-SAM and Grounding DINO T. Specifically, we randomly sample half of the data from EVF-SAM outputs and the other half from Grounding DINO T to form a balanced mixture.

\section{PiTER protocol Analysis}
PiTER evaluation protocol removes intermediate reasoning, tool-calling, and iterative zooming, which are typically integral components of TiVR methods. However, this design choice is purposeful in the context of our experimental setup. All evaluated Tool-use models are tuned from the same pretrained backbone, Qwen2.5-VL-7B, meaning that their foundational visual–language capabilities and architectural priors are identical. The objective of PiTER is therefore not to compare full agentic pipelines, but to isolate the incremental refinement ability contributed by each fine-tuning strategy beyond the shared base model. Allowing each method to operate with its native multi-step interaction style would introduce confounding factors, as performance would reflect differences in external scaffolding rather than the refinement competence learned by the model itself. By enforcing a single-stage, tool-injected inference format, PiTER normalizes the interaction interface across all methods and evaluates their intrinsic ability to utilize imperfect external tool outputs. This uniform setting ensures that any performance differences stem from the models’ learned refinement mechanisms rather than disparities in procedural reasoning frameworks.

\section{More Case Study}
Figure \ref{sup_vis_1} presents additional examples where our VG-Refiner successfully corrects tool failures. Most of these cases originate from the non-finetuned Grounding DINO T, which exhibits relatively weak capability in modeling spatial relationships and object–attribute associations.

%% file: main.bib
@String(AAAI = {AAAI})

@article{vtool-r1,
  title={VTool-R1: VLMs Learn to Think with Images via Reinforcement Learning on Multimodal Tool Use},
  author={Wu, Mingyuan and Yang, Jingcheng and Jiang, Jize and Li, Meitang and Yan, Kaizhuo and Yu, Hanchao and Zhang, Minjia and Zhai, Chengxiang and Nahrstedt, Klara},
  journal={arXiv preprint arXiv:2505.19255},
  year={2025}
}

@article{dianjinocr,
  title={Dianjin-ocr-r1: Enhancing ocr capabilities via a reasoning-and-tool interleaved vision-language model},
  author={Chen, Qian and Zhang, Xianyin and Guo, Lifan and Chen, Feng and Zhang, Chi},
  journal={arXiv preprint arXiv:2508.13238},
  year={2025}
}

@article{openthinkimg,
  title={Openthinkimg: Learning to think with images via visual tool reinforcement learning},
  author={Su, Zhaochen and Li, Linjie and Song, Mingyang and Hao, Yunzhuo and Yang, Zhengyuan and Zhang, Jun and Chen, Guanjie and Gu, Jiawei and Li, Juntao and Qu, Xiaoye and others},
  journal={arXiv preprint arXiv:2505.08617},
  year={2025}
}

@article{deepeyes,
  title={Deepeyes: Progressive visual analytics for designing deep neural networks},
  author={Pezzotti, Nicola and H{\"o}llt, Thomas and Van Gemert, Jan and Lelieveldt, Boudewijn PF and Eisemann, Elmar and Vilanova, Anna},
  journal={IEEE transactions on visualization and computer graphics},
  volume={24},
  number={1},
  pages={98--108},
  year={2017},
  publisher={IEEE}
}

@article{search-r1,
  title={Search-r1: Training llms to reason and leverage search engines with reinforcement learning},
  author={Jin, Bowen and Zeng, Hansi and Yue, Zhenrui and Yoon, Jinsung and Arik, Sercan and Wang, Dong and Zamani, Hamed and Han, Jiawei},
  journal={arXiv preprint arXiv:2503.09516},
  year={2025}
}

@article{thinking-with-videos,
  title={Thinking with videos: Multimodal tool-augmented reinforcement learning for long video reasoning},
  author={Zhang, Haoji and Gu, Xin and Li, Jiawen and Ma, Chixiang and Bai, Sule and Zhang, Chubin and Zhang, Bowen and Zhou, Zhichao and He, Dongliang and Tang, Yansong},
  journal={arXiv preprint arXiv:2508.04416},
  year={2025}
}

@article{ponder,
  title={Ponder \& press: Advancing visual gui agent towards general computer control},
  author={Wang, Yiqin and Zhang, Haoji and Tian, Jingqi and Tang, Yansong},
  journal={arXiv preprint arXiv:2412.01268},
  year={2024}
}

@article{seg-zero,
  title={Seg-zero: Reasoning-chain guided segmentation via cognitive reinforcement},
  author={Liu, Yuqi and Peng, Bohao and Zhong, Zhisheng and Yue, Zihao and Lu, Fanbin and Yu, Bei and Jia, Jiaya},
  journal={arXiv preprint arXiv:2503.06520},
  year={2025}
}

@article{univg,
  title={Univg-r1: Reasoning guided universal visual grounding with reinforcement learning},
  author={Bai, Sule and Li, Mingxing and Liu, Yong and Tang, Jing and Zhang, Haoji and Sun, Lei and Chu, Xiangxiang and Tang, Yansong},
  journal={arXiv preprint arXiv:2505.14231},
  year={2025}
}

@article{deepseek-r1,
  title={Deepseek-r1 incentivizes reasoning in llms through reinforcement learning},
  author={Guo, Daya and Yang, Dejian and Zhang, Haowei and Song, Junxiao and Wang, Peiyi and Zhu, Qihao and Xu, Runxin and Zhang, Ruoyu and Ma, Shirong and Bi, Xiao and others},
  journal={Nature},
  volume={645},
  number={8081},
  pages={633--638},
  year={2025},
  publisher={Nature Publishing Group UK London}
}

@article{visual-rft,
  title={Visual-rft: Visual reinforcement fine-tuning},
  author={Liu, Ziyu and Sun, Zeyi and Zang, Yuhang and Dong, Xiaoyi and Cao, Yuhang and Duan, Haodong and Lin, Dahua and Wang, Jiaqi},
  journal={arXiv preprint arXiv:2503.01785},
  year={2025}
}

@article{revpt,
  title={Reinforced visual perception with tools},
  author={Zhou, Zetong and Chen, Dongping and Ma, Zixian and Hu, Zhihan and Fu, Mingyang and Wang, Sinan and Wan, Yao and Zhao, Zhou and Krishna, Ranjay},
  journal={arXiv preprint arXiv:2509.01656},
  year={2025}
}

@article{visual-arft,
  title={Visual Agentic Reinforcement Fine-Tuning},
  author={Liu, Ziyu and Zang, Yuhang and Zou, Yushan and Liang, Zijian and Dong, Xiaoyi and Cao, Yuhang and Duan, Haodong and Lin, Dahua and Wang, Jiaqi},
  journal={arXiv preprint arXiv:2505.14246},
  year={2025}
}

@article{thyme,
  title={Thyme: Think beyond images},
  author={Zhang, Yi-Fan and Lu, Xingyu and Yin, Shukang and Fu, Chaoyou and Chen, Wei and Hu, Xiao and Wen, Bin and Jiang, Kaiyu and Liu, Changyi and Zhang, Tianke and others},
  journal={arXiv preprint arXiv:2508.11630},
  year={2025}
}

@article{VILASR,
  title={Reinforcing spatial reasoning in vision-language models with interwoven thinking and visual drawing},
  author={Wu, Junfei and Guan, Jian and Feng, Kaituo and Liu, Qiang and Wu, Shu and Wang, Liang and Wu, Wei and Tan, Tieniu},
  journal={arXiv preprint arXiv:2506.09965},
  year={2025}
}

@article{survey-agent,
  title={A Survey on Agentic Multimodal Large Language Models},
  author={Yao, Huanjin and Zhang, Ruifei and Huang, Jiaxing and Zhang, Jingyi and Wang, Yibo and Fang, Bo and Zhu, Ruolin and Jing, Yongcheng and Liu, Shunyu and Li, Guanbin and others},
  journal={arXiv preprint arXiv:2510.10991},
  year={2025}
}

@inproceedings{grounding-dino,
  title={Grounding dino: Marrying dino with grounded pre-training for open-set object detection},
  author={Liu, Shilong and Zeng, Zhaoyang and Ren, Tianhe and Li, Feng and Zhang, Hao and Yang, Jie and Jiang, Qing and Li, Chunyuan and Yang, Jianwei and Su, Hang and others},
  booktitle={European conference on computer vision},
  pages={38--55},
  year={2024},
  organization={Springer}
}

@article{grounding-dino1.5,
  title={Grounding dino 1.5: Advance the" edge" of open-set object detection},
  author={Ren, Tianhe and Jiang, Qing and Liu, Shilong and Zeng, Zhaoyang and Liu, Wenlong and Gao, Han and Huang, Hongjie and Ma, Zhengyu and Jiang, Xiaoke and Chen, Yihao and others},
  journal={arXiv preprint arXiv:2405.10300},
  year={2024}
}

@inproceedings{depth-anything,
  title={Depth anything: Unleashing the power of large-scale unlabeled data},
  author={Yang, Lihe and Kang, Bingyi and Huang, Zilong and Xu, Xiaogang and Feng, Jiashi and Zhao, Hengshuang},
  booktitle={Proceedings of the IEEE/CVF conference on computer vision and pattern recognition},
  pages={10371--10381},
  year={2024}
}

@article{depth-anythingv2,
  title={Depth anything v2},
  author={Yang, Lihe and Kang, Bingyi and Huang, Zilong and Zhao, Zhen and Xu, Xiaogang and Feng, Jiashi and Zhao, Hengshuang},
  journal={Advances in Neural Information Processing Systems},
  volume={37},
  pages={21875--21911},
  year={2024}
}

@article{rec-survey,
  title={Referring expression comprehension: A survey of methods and datasets},
  author={Qiao, Yanyuan and Deng, Chaorui and Wu, Qi},
  journal={IEEE Transactions on Multimedia},
  volume={23},
  pages={4426--4440},
  year={2020},
  publisher={IEEE}
}

@inproceedings{lavt,
  title={Lavt: Language-aware vision transformer for referring image segmentation},
  author={Yang, Zhao and Wang, Jiaqi and Tang, Yansong and Chen, Kai and Zhao, Hengshuang and Torr, Philip HS},
  booktitle={Proceedings of the IEEE/CVF conference on computer vision and pattern recognition},
  pages={18155--18165},
  year={2022}
}

@inproceedings{llava-grounding,
  title={Llava-grounding: Grounded visual chat with large multimodal models},
  author={Zhang, Hao and Li, Hongyang and Li, Feng and Ren, Tianhe and Zou, Xueyan and Liu, Shilong and Huang, Shijia and Gao, Jianfeng and Leizhang and Li, Chunyuan and others},
  booktitle={European Conference on Computer Vision},
  pages={19--35},
  year={2024},
  organization={Springer}
}

@inproceedings{lisa,
  title={Lisa: Reasoning segmentation via large language model},
  author={Lai, Xin and Tian, Zhuotao and Chen, Yukang and Li, Yanwei and Yuan, Yuhui and Liu, Shu and Jia, Jiaya},
  booktitle={Proceedings of the IEEE/CVF Conference on Computer Vision and Pattern Recognition},
  pages={9579--9589},
  year={2024}
}

@inproceedings{referring-counting,
  title={Referring expression counting},
  author={Dai, Siyang and Liu, Jun and Cheung, Ngai-Man},
  booktitle={Proceedings of the IEEE/CVF Conference on Computer Vision and Pattern Recognition},
  pages={16985--16995},
  year={2024}
}

@article{qwen2.5VL,
  title={Qwen2. 5-vl technical report},
  author={Bai, Shuai and Chen, Keqin and Liu, Xuejing and Wang, Jialin and Ge, Wenbin and Song, Sibo and Dang, Kai and Wang, Peng and Wang, Shijie and Tang, Jun and others},
  journal={arXiv preprint arXiv:2502.13923},
  year={2025}
}

@article{mini_o3,
  title={Mini-o3: Scaling up reasoning patterns and interaction turns for visual search},
  author={Lai, Xin and Li, Junyi and Li, Wei and Liu, Tao and Li, Tianjian and Zhao, Hengshuang},
  journal={arXiv preprint arXiv:2509.07969},
  year={2025}
}

@article{pixel_reasoner,
  title={Pixel reasoner: Incentivizing pixel-space reasoning with curiosity-driven reinforcement learning},
  author={Su, Alex and Wang, Haozhe and Ren, Weiming and Lin, Fangzhen and Chen, Wenhu},
  journal={arXiv preprint arXiv:2505.15966},
  year={2025}
}

@article{thinking_withimage_survey,
  title={Thinking with images for multimodal reasoning: Foundations, methods, and future frontiers},
  author={Su, Zhaochen and Xia, Peng and Guo, Hangyu and Liu, Zhenhua and Ma, Yan and Qu, Xiaoye and Liu, Jiaqi and Li, Yanshu and Zeng, Kaide and Yang, Zhengyuan and others},
  journal={arXiv preprint arXiv:2506.23918},
  year={2025}
}

@article{active_o3,
  title={Active-O3: Empowering Multimodal Large Language Models with Active Perception via GRPO},
  author={Zhu, Muzhi and Zhong, Hao and Zhao, Canyu and Du, Zongze and Huang, Zheng and Liu, Mingyu and Chen, Hao and Zou, Cheng and Chen, Jingdong and Yang, Ming and others},
  journal={arXiv preprint arXiv:2505.21457},
  year={2025}
}

@article{simple_o3,
  title={Simple o3: Towards interleaved vision-language reasoning},
  author={Wang, Ye and Chen, Qianglong and Li, Zejun and Wang, Siyuan and Guo, Shijie and Zhang, Zhirui and Wei, Zhongyu},
  journal={arXiv preprint arXiv:2508.12109},
  year={2025}
}

@inproceedings{sam2-love,
  title={SAM2-LOVE: Segment Anything Model 2 in Language-aided Audio-Visual Scenes},
  author={Wang, Yuji and Xu, Haoran and Liu, Yong and Li, Jiaze and Tang, Yansong},
  booktitle={Proceedings of the Computer Vision and Pattern Recognition Conference},
  pages={28932--28941},
  year={2025}
}

@article{flash-vstream,
  title={Flash-VStream: Efficient Real-Time Understanding for Long Video Streams},
  author={Zhang, Haoji and Wang, Yiqin and Tang, Yansong and Liu, Yong and Feng, Jiashi and Jin, Xiaojie},
  journal={arXiv preprint arXiv:2506.23825},
  year={2025}
}

@article{rex_omni,
  title={Detect Anything via Next Point Prediction},
  author={Jiang, Qing and Huo, Junan and Chen, Xingyu and Xiong, Yuda and Zeng, Zhaoyang and Chen, Yihao and Ren, Tianhe and Yu, Junzhi and Zhang, Lei},
  journal={arXiv preprint arXiv:2510.12798},
  year={2025}
}

@inproceedings{iterprime,
  title={Iterprime: Zero-shot referring image segmentation with iterative grad-cam refinement and primary word emphasis},
  author={Wang, Yuji and Ni, Jingchen and Liu, Yong and Yuan, Chun and Tang, Yansong},
  booktitle={Proceedings of the AAAI Conference on Artificial Intelligence},
  volume={39},
  number={8},
  pages={8159--8168},
  year={2025}
}

@article{soc,
  title={Semantic-Assisted Object Clustering for Multi-Modal Referring Video Segmentation},
  author={Liu, Yong and Luo, Zhuoyan and Xiao, Yicheng and Wang, Yitong and Li, Shuyan and Li, Xiu and Yang, Yujiu and Tang, Yansong},
  journal={IEEE Transactions on Pattern Analysis and Machine Intelligence},
  year={2025},
  publisher={IEEE}
}

@inproceedings{unilseg,
  title={Universal segmentation at arbitrary granularity with language instruction},
  author={Liu, Yong and Zhang, Cairong and Wang, Yitong and Wang, Jiahao and Yang, Yujiu and Tang, Yansong},
  booktitle={Proceedings of the IEEE/CVF Conference on Computer Vision and Pattern Recognition},
  pages={3459--3469},
  year={2024}
}

@article{llava,
  title={Visual instruction tuning},
  author={Liu, Haotian and Li, Chunyuan and Wu, Qingyang and Lee, Yong Jae},
  journal={Advances in neural information processing systems},
  volume={36},
  pages={34892--34916},
  year={2023}
}

@article{qwen2,
  title={Qwen2-vl: Enhancing vision-language model's perception of the world at any resolution},
  author={Wang, Peng and Bai, Shuai and Tan, Sinan and Wang, Shijie and Fan, Zhihao and Bai, Jinze and Chen, Keqin and Liu, Xuejing and Wang, Jialin and Ge, Wenbin and others},
  journal={arXiv preprint arXiv:2409.12191},
  year={2024}
}

@article{qwen3_omni,
  title={Qwen3-omni technical report},
  author={Xu, Jin and Guo, Zhifang and Hu, Hangrui and Chu, Yunfei and Wang, Xiong and He, Jinzheng and Wang, Yuxuan and Shi, Xian and He, Ting and Zhu, Xinfa and others},
  journal={arXiv preprint arXiv:2509.17765},
  year={2025}
}

@article{groundinggpt,
  title={Groundinggpt: Language enhanced multi-modal grounding model},
  author={Li, Zhaowei and Xu, Qi and Zhang, Dong and Song, Hang and Cai, Yiqing and Qi, Qi and Zhou, Ran and Pan, Junting and Li, Zefeng and Vu, Van Tu and others},
  journal={arXiv preprint arXiv:2401.06071},
  year={2024}
}

@article{vlm-r1,
  title={Vlm-r1: A stable and generalizable r1-style large vision-language model},
  author={Shen, Haozhan and Liu, Peng and Li, Jingcheng and Fang, Chunxin and Ma, Yibo and Liao, Jiajia and Shen, Qiaoli and Zhang, Zilun and Zhao, Kangjia and Zhang, Qianqian and others},
  journal={arXiv preprint arXiv:2504.07615},
  year={2025}
}

@article{ground-r1,
  title={Ground-R1: Incentivizing Grounded Visual Reasoning via Reinforcement Learning},
  author={Cao, Meng and Zhao, Haoze and Zhang, Can and Chang, Xiaojun and Reid, Ian and Liang, Xiaodan},
  journal={arXiv preprint arXiv:2505.20272},
  year={2025}
}

@article{visionreasoner,
  title={VisionReasoner: Unified Visual Perception and Reasoning via Reinforcement Learning},
  author={Liu, Yuqi and Qu, Tianyuan and Zhong, Zhisheng and Peng, Bohao and Liu, Shu and Yu, Bei and Jia, Jiaya},
  journal={arXiv preprint arXiv:2505.12081},
  year={2025}
}

@article{rex_thinker,
  title={Rex-Thinker: Grounded Object Referring via Chain-of-Thought Reasoning},
  author={Jiang, Qing and Chen, Xingyu and Zeng, Zhaoyang and Yu, Junzhi and Zhang, Lei},
  journal={arXiv preprint arXiv:2506.04034},
  year={2025}
}

@inproceedings{rex_seek,
  title={Referring to any person},
  author={Jiang, Qing and Wu, Lin and Zeng, Zhaoyang and Ren, Tianhe and Xiong, Yuda and Chen, Yihao and Qin, Liu and Zhang, Lei},
  booktitle={Proceedings of the IEEE/CVF International Conference on Computer Vision},
  pages={21667--21678},
  year={2025}
}

@inproceedings{refcoco,
  title={Modeling context in referring expressions},
  author={Yu, Licheng and Poirson, Patrick and Yang, Shan and Berg, Alexander C and Berg, Tamara L},
  booktitle={European conference on computer vision},
  pages={69--85},
  year={2016},
  organization={Springer}
}

@article{evf-sam,
  title={Evf-sam: Early vision-language fusion for text-prompted segment anything model},
  author={Zhang, Yuxuan and Cheng, Tianheng and Zhu, Lianghui and Hu, Rui and Liu, Lei and Liu, Heng and Ran, Longjin and Chen, Xiaoxin and Liu, Wenyu and Wang, Xinggang},
  journal={arXiv preprint arXiv:2406.20076},
  year={2024}
}

@article{InstanceVG,
  title={Improving generalized visual grounding with instance-aware joint learning},
  author={Dai, Ming and Cheng, Wenxuan and Liu, Jiang-Jiang and Yang, Lingfeng and Feng, Zhenhua and Yang, Wankou and Wang, Jingdong},
  journal={IEEE Transactions on Pattern Analysis and Machine Intelligence},
  year={2025},
  publisher={IEEE}
}

@inproceedings{beit_3,
  title={Image as a foreign language: Beit pretraining for vision and vision-language tasks},
  author={Wang, Wenhui and Bao, Hangbo and Dong, Li and Bjorck, Johan and Peng, Zhiliang and Liu, Qiang and Aggarwal, Kriti and Mohammed, Owais Khan and Singhal, Saksham and Som, Subhojit and others},
  booktitle={Proceedings of the IEEE/CVF Conference on Computer Vision and Pattern Recognition},
  pages={19175--19186},
  year={2023}
}

@inproceedings{mmbench,
  title={Mmbench: Is your multi-modal model an all-around player?},
  author={Liu, Yuan and Duan, Haodong and Zhang, Yuanhan and Li, Bo and Zhang, Songyang and Zhao, Wangbo and Yuan, Yike and Wang, Jiaqi and He, Conghui and Liu, Ziwei and others},
  booktitle={European conference on computer vision},
  pages={216--233},
  year={2024},
  organization={Springer}
}

@article{mmstar,
  title={Are we on the right way for evaluating large vision-language models?},
  author={Chen, Lin and Li, Jinsong and Dong, Xiaoyi and Zhang, Pan and Zang, Yuhang and Chen, Zehui and Duan, Haodong and Wang, Jiaqi and Qiao, Yu and Lin, Dahua and others},
  journal={Advances in Neural Information Processing Systems},
  volume={37},
  pages={27056--27087},
  year={2024}
}

@article{ocrbench,
  title={Ocrbench: on the hidden mystery of ocr in large multimodal models},
  author={Liu, Yuliang and Li, Zhang and Huang, Mingxin and Yang, Biao and Yu, Wenwen and Li, Chunyuan and Yin, Xu-Cheng and Liu, Cheng-Lin and Jin, Lianwen and Bai, Xiang},
  journal={Science China Information Sciences},
  volume={67},
  number={12},
  pages={220102},
  year={2024},
  publisher={Springer}
}

@inproceedings{chartqa,
  title={Chartqa: A benchmark for question answering about charts with visual and logical reasoning},
  author={Masry, Ahmed and Do, Xuan Long and Tan, Jia Qing and Joty, Shafiq and Hoque, Enamul},
  booktitle={Findings of the association for computational linguistics: ACL 2022},
  pages={2263--2279},
  year={2022}
}

@misc{realwordqa,
  title        = {Grok-1.5 Vision Preview},
  author       = {xAI},
  year         = {2024},
  month        = apr,
  howpublished = {\url{https://x.ai/news/grok-1.5v}}
}

@inproceedings{verl,
  title={Hybridflow: A flexible and efficient rlhf framework},
  author={Sheng, Guangming and Zhang, Chi and Ye, Zilingfeng and Wu, Xibin and Zhang, Wang and Zhang, Ru and Peng, Yanghua and Lin, Haibin and Wu, Chuan},
  booktitle={Proceedings of the Twentieth European Conference on Computer Systems},
  pages={1279--1297},
  year={2025}
}

@inproceedings{vllm,
  title={Efficient memory management for large language model serving with pagedattention},
  author={Kwon, Woosuk and Li, Zhuohan and Zhuang, Siyuan and Sheng, Ying and Zheng, Lianmin and Yu, Cody Hao and Gonzalez, Joseph and Zhang, Hao and Stoica, Ion},
  booktitle={Proceedings of the 29th symposium on operating systems principles},
  pages={611--626},
  year={2023}
}

@article{traceable,
  title={Traceable evidence enhanced visual grounded reasoning: Evaluation and methodology},
  author={Wang, Haochen and Li, Xiangtai and Huang, Zilong and Wang, Anran and Wang, Jiacong and Zhang, Tao and Zheng, Jiani and Bai, Sule and Kang, Zijian and Feng, Jiashi and others},
  journal={arXiv preprint arXiv:2507.07999},
  year={2025}
}

@article{cogcom,
  title={Cogcom: Train large vision-language models diving into details through chain of manipulations},
  author={Qi, Ji and Ding, Ming and Wang, Weihan and Bai, Yushi and Lv, Qingsong and Hong, Wenyi and Xu, Bin and Hou, Lei and Li, Juanzi and Dong, Yuxiao and others},
  year={2024}
}

@article{cogvlm,
  title={Cogvlm: Visual expert for pretrained language models},
  author={Wang, Weihan and Lv, Qingsong and Yu, Wenmeng and Hong, Wenyi and Qi, Ji and Wang, Yan and Ji, Junhui and Yang, Zhuoyi and Zhao, Lei and XiXuan, Song and others},
  journal={Advances in Neural Information Processing Systems},
  volume={37},
  pages={121475--121499},
  year={2024}
}

@article{vitron,
  title={Vitron: A unified pixel-level vision llm for understanding, generating, segmenting, editing},
  author={Fei, Hao and Wu, Shengqiong and Zhang, Hanwang and Chua, Tat-Seng and Yan, Shuicheng},
  journal={Advances in neural information processing systems},
  volume={37},
  pages={57207--57239},
  year={2024}
}

@article{unipixel,
  title={Unipixel: Unified object referring and segmentation for pixel-level visual reasoning},
  author={Liu, Ye and Ma, Zongyang and Pu, Junfu and Qi, Zhongang and Wu, Yang and Shan, Ying and Chen, Chang Wen},
  journal={arXiv preprint arXiv:2509.18094},
  year={2025}
}

@inproceedings{uninext,
  title={Universal instance perception as object discovery and retrieval},
  author={Yan, Bin and Jiang, Yi and Wu, Jiannan and Wang, Dong and Luo, Ping and Yuan, Zehuan and Lu, Huchuan},
  booktitle={Proceedings of the IEEE/CVF Conference on Computer Vision and Pattern Recognition},
  pages={15325--15336},
  year={2023}
}

@inproceedings{llmdet,
  title={Llmdet: Learning strong open-vocabulary object detectors under the supervision of large language models},
  author={Fu, Shenghao and Yang, Qize and Mo, Qijie and Yan, Junkai and Wei, Xihan and Meng, Jingke and Xie, Xiaohua and Zheng, Wei-Shi},
  booktitle={Proceedings of the Computer Vision and Pattern Recognition Conference},
  pages={14987--14997},
  year={2025}
}
